\documentclass[preprint,12pt]{elsarticle}

\usepackage{amssymb}
\usepackage{float}
\usepackage{comment}

\usepackage{amsfonts,amssymb,amsmath,graphicx,dsfont}
\usepackage{amsthm}
\usepackage[english]{babel}

\usepackage{caption}
\usepackage{subcaption}

\newcommand{\ra}{\renewcommand{\arraystretch}}
\usepackage{lineno}
\usepackage[dvipsnames]{xcolor}
\usepackage{booktabs}
\usepackage{mathtools} 
\usepackage{textcomp, gensymb}

\usepackage[utf8]{inputenc}
\usepackage{url}

\journal{Applied Energy}

\begin{document}

\begin{frontmatter}



\title{Electricity Demand Forecasting with Hybrid Statistical and Machine Learning Algorithms: Case Study of Ukraine}
\author[add2]{T. Gonz\'alez Grand\'on } 
\ead{tatiana.grandon@uni-flensburg.de}

\author[add3]{J. Schwenzer}
\ead{johannes.m.schwenzer@gmail.com}

\author[add4]{T. Steens} 
\ead{thomas.steens@dlr.de}

\author{J. Breuing} 
\ead{julia@breuing.eu}

\address[add2]{Europa-Universit\"{a}t Flensburg, Munketoft 3, 24937 Flensburg }
\address[add3]{Europa-Universit\"{a}t Viadrina, Grosse Scharrnst. 59, Frankfurt, Germany}
\address[add4]{DLR Institute of Networked Energy Systems
Carl-von-Ossietzky-Str. 15, Germany}

\begin{abstract} 
This article presents a novel hybrid approach using statistics and machine learning to forecast the national demand of electricity. As investment and operation of future energy systems require long-term electricity demand forecasts with hourly resolution, our mathematical model fills a gap in energy forecasting. The proposed methodology was constructed using hourly data from Ukraine's electricity consumption ranging from 2013 to 2020.
To this end, we analysed the underlying structure of the hourly, daily and yearly time series of electricity consumption.
The long-term yearly trend is evaluated using macroeconomic regression analysis. The mid-term model integrates temperature and calendar regressors to describe the underlying structure, and combines ARIMA and LSTM ``black-box'' pattern-based approaches to describe the error term. The short-term model captures the hourly seasonality through calendar regressors and multiple ARMA models for the residual. 
Results show that the best forecasting model is composed by combining multiple regression models and a LSTM hybrid model for residual prediction.  Our hybrid model is
very effective at forecasting long-term electricity consumption on an hourly resolution. In two years of out-of-sample forecasts with 17520 timesteps, it is shown to be within 96.83 \% accuracy. 
 
\end{abstract}

\begin{keyword}
 national electricity demand \sep forecasting \sep ARIMA \sep LSTM. 


\end{keyword}

\end{frontmatter}

\nolinenumbers

\section{Introduction}\label{Intro}

Electricity plays an important role in the growth and development of nations. It has been proven that whether in developing or industrialized geographies, electricity consumption plays a pertinent role in the process of economic growth and in the low-carbon transformation of production \cite{EcoGrowth1,eco2,eco3}. Therefore, accurate forecasts of the load demand are important for strategic decision making in any nation. It is a field where intensive research is being conducted, as even minimal improvements in forecasting accuracy are worth millions of Euros \cite{Hobbs1,million}. For example, it was estimated in \cite{Bunn} that a mere 1 \% reduction in the forecast error resulted in a decrease of \textsterling 10 million in operating costs per year for an electric utility in the United Kingdom. Despite this, the existing body of literature on demand forecasting, still remains a challenge for scientists and decision-makers \cite{Carvallo, Lindberg}. 

Forecasts of different time-horizons and different accuracy are required for the power system operation and investment in power generation renewal. Based on the time horizon and time resolution, forecasting is assorted into short-term, medium-term and long-term. Short-term load forecasting (STLF) is particularly essential for day-ahead or intra-day power market operation. The time horizon for STLF is one-day or a week and the time resolution is on hourly or minute-scale. Medium-term load forecasting (MTLF) is mostly concerned with fuel import and maintenance scheduling. The MTLF time horizon is based on a monthly scale \cite{booksmart}, while the time resolution is usually on a daily scale. Long-term load forecasting (LTLF) is important in power system planning (generation, transmission and distribution) and investment in new generation units. LTLF time horizon ranges from a few years to ten years with monthly granularity \cite{LT1}. However, today's current energy transition, most decision processes and scenario analyses - such as the integration of high volatile renewable energies, end user flexibility, large scale electric mobility or other Power-to-X technologies - require one model with multiple time horizons. With an increasing shift to renewable energy generation as well as a higher penetration of electricity into the primary energy consumption, the need for accurate LTLF in fine resolution is crucial. The authors therefore fill a knowledge gap on an accurate model for LTLF in an hourly resolution.


\subsection{Literature Review}
Load forecasting models can be, in general, divided into two large classes - namely classical statistic techniques (CST) and modern statistical learning based approaches from machine learning (ML). The most widely used classical statistic methods are multiple regression \cite{Pielow2012,Turkey2017,reg2,Greekreg, Greekreg2, reg3, reg4}, and autoregressive structures \cite{Ghana,China,Filik2011,arimaANN, Brazil}.  Multiple regression and time series autoregressive approaches demand an \textit{a priori} definition of the structure of the model. The regression models studied are LTLF with monthly granularity and have reached a forecast accuracy with a mean average percent error (MAPE) of $2.6 \%$. ARIMA models are based on the assumption that the data has an internal structure in the form of trend, seasonality, auto-correlation and non-stationarity \cite{Feinberg2005}. The ARIMA forecast models reviewed are STLF and have an average MAPE value of $3.9 \%$.

On the contrary, machine learning models such as neural networks are trained by historical data, such that the structure of the model is defined on an \textit{a posteriori} basis \cite{MLreview}. The most widely used ML methods for electricity forecasting are Artificial Neural Networks (ANN) \cite{ANN1,ANN2, Iran, ANN3, IndiaANN} and  Recurrent Neural Networks (RNN) \cite{RNN1,RNN2,RNN3}. The former approach reached a forecasting accuracy with an average MAPE of $3.7 \%$, while the latter RNN approach reached an average MAPE value of $3.6 \%$. 

Both CST and ML aim at improving forecasting accuracy by minimizing a defined loss function. The difference lies in how the minimization of an error function between the predicted and actual value is performed. The CST use linear processes, while ML uses nonlinear algorithms. Thus, CST methods are usually less computationally complex \cite{Makridakis} but ML methods are becoming increasingly popular since they also account for the nonlinear behaviour of electricity demand as well. Nonetheless, ML approaches are mostly used for short-term load predictions. Until now there is limited research in their capability of long-term forecasting \cite{Ghods2008}, hence the focus of this paper. Also, ML methods act as a black-box, so no specific relationship between load and other variables can be incorporated beforehand. Some articles, like \citet{arimaANN}, compare the two approaches and conclude that the ARIMA model has a lower prediction error with a MAPE of 15 \% than the ANN model. Also, \citet{Taiwan} compare ARIMA and ANN models to predict energy consumption in Taiwan. They conclude that a single variable ARIMA model performs better. However, the aforementioned articles evaluate the different forecasting methods separately, as opposed to combining them. 

Methodologies for forecasting the national electricity demand and its hourly profile, are generally based on extrapolating current load profiles \cite{Lindberg,Lb2,Boro}. Some authors like \citet{Andersen2013} have introduced a model with CST for LTLF with hourly resolution. However, their model requires hourly metering of individual customers in a nation. Such a large amount of data coming from individual metering hardware is usually not available in most industrialized nations, much less in developing countries. Our study is inspired on the work of \citet{Pielow2012} and 
\citet{Sotiropoulos2013}. The former models STLF  by employing Fourier series and scales it with long-term growth rates based on a linear regression. \citet{Sotiropoulos2013} present a medium-term model for daily averages and a short-term model for hourly residuals. Both time horizons consist of a deterministic part with exogenous regressors and a stochastic part modelled by autoregressive terms.
Although we were inspired by this methodology, we divided our model in three horizons instead of two (to include LSTF).  Moreover, to improve forecasting accuracy, while still having an understanding of the drivers influencing different time horizons, we combined CST and ML approaches. 
Currently, similar to this present article, combinations of CST and ML models are being studied to provide better forecasting accuracy and delude deficiencies from each approach. The methodology of forecast combination, called ensemble or hybrid forecasting, is a suitable tool that has been successfully applied \cite{Bunn2, Clemen}. \citet{Taiwan2} used a hybrid ARIMA-NN model to predict wind power generation in Taiwan, however this method is only suitable for short-term prediction. \citet{Wang} combine a seasonal autoregressive intergated moving average (SARIMA) and NN models by an unequal weighting using a variance-covariance approach. \citet{online} present a combined model, which improves a forecast particularly on special days. For example, the results from the public holidays in France showed an average MAPE of 0.863 \%. The results of these studies show the potential for hybrid models, yet they are currently only used for short-term load forecasting. 

 To our knowledge, there is a lack of studies that discuss the use of hybrid methods for long-term load forecasting with hourly resolution. Some authors have already addressed the topic of long-term modelling on an hourly basis, such as \cite{Andersen2013, Pielow2012, Sotiropoulos2013, Filik2011}. The downside of these studies is that there is only one choice of models and they are all based on CST. 
 
 The main aim of this article is to present a novel forecast hybrid method, namely a CST-RNN method for predicting the overall electricity demand of a nation on a long-term hourly resolution. As a case study, we evaluate the data of Ukraine from 2013 to 2020 from \cite{Ukrenergo2020a}. 

The main contributions of this article are: 
\begin{enumerate}
 \item A long-term prediction model with hourly resolution is proposed following a classical approach, with concepts stemming from regression analysis and ARIMA models.
    \item A hybrid CST-LSTM model is presented for forecasting errors to correct CST and improve forecast accuracy.
    \item Comparison and drawbacks of each model are shown using four statistical parameters.  
    \item A detailed methodology for the  computation, including an open-source code, which can be fully automatized and reproduced for future load prediction in other countries.

\end{enumerate}

\noindent The remaining paper is organized as follows. Section 2 contains an overview of Ukraine's electricity sector. The model decomposition and the statistical description for the respective long, medium and short-term components are presented in Section 3. In Section 4, the performance of the algorithm is presented and tested by forecasting electricity demand in 2019 and 2020. 
 Section 5 concludes the research and describes future research initiatives. 

\section{Overview of Ukraine's Electricity Demand} \label{DM}

Ukraine is a developing country, which is currently undergoing a political and economic crisis. 
Changes of electricity consumption in the country play an increasingly critical role in its dilemma of energy security. The Ukrainian electricity sector has been reformed and a connection of the Ukrainian grid to the continental European grid was intended and is now in progress \cite{swp}. 
 
\begin{figure}[ht]
\centering
  \includegraphics[width=0.55\linewidth]{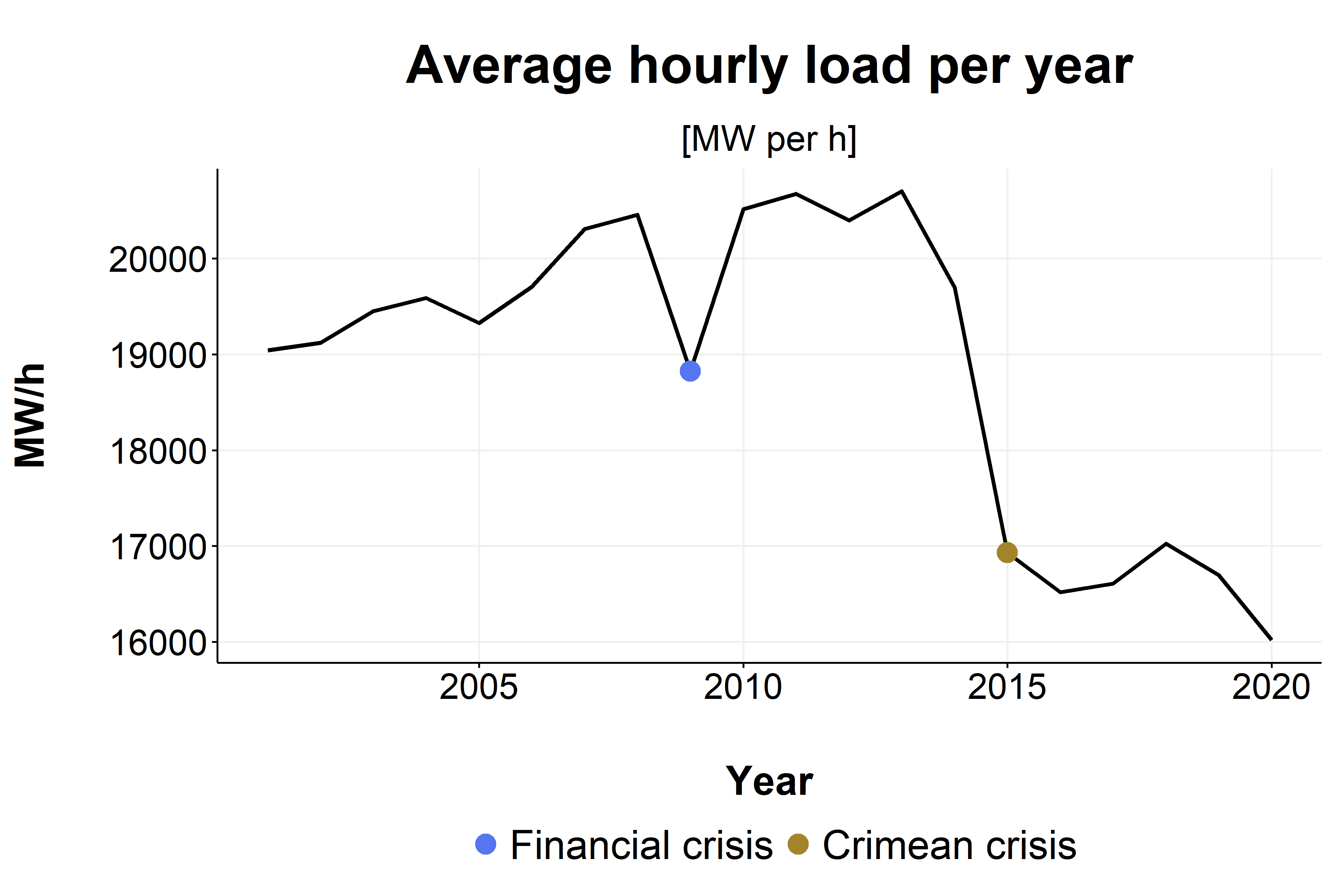}
  \caption{Ukraine's yearly average hourly load over the course of the years 2000 to 2020. }\label{fig:LongTermOverview}
\end{figure}

Electricity consumption is affected by macroeconomic factors, which influence the load series trend and the yearly periodicity. Figure \ref{fig:LongTermOverview} illustrates Ukraine's electricity demand for the period of 2001 to 2020. The upward trend is interrupted in 2009, in the wake of the global financial crisis. Additionally, the Crimean crisis has led to a substantial drop of 30 TWh in the total annual electricity demand. This decline can be partly explained by the annexation of Crimea to Russia, as the peninsula is no longer connected to the Ukrainian grid. Crimean demand made up for about $5\%$ of the pre-crisis level \cite{Urkstat2019}, so only 10 TWh can be attributed to this. The Donbass region was the second largest contributor to electricity demand before the crisis. In 2014, this region accounted for $19\%$ of Ukraine's load. Its share decreased to $9\%$ in 2018 \cite{Ukrenergo2020a} due to the economic depression \cite{Ginsborg2019}. This highlights three macroeconomic variables that commonly influence the electricity demand time series: population, GDP, and GDP growth rate.   

\begin{figure}[H]
    \centering
    \includegraphics[width=0.70\textwidth]{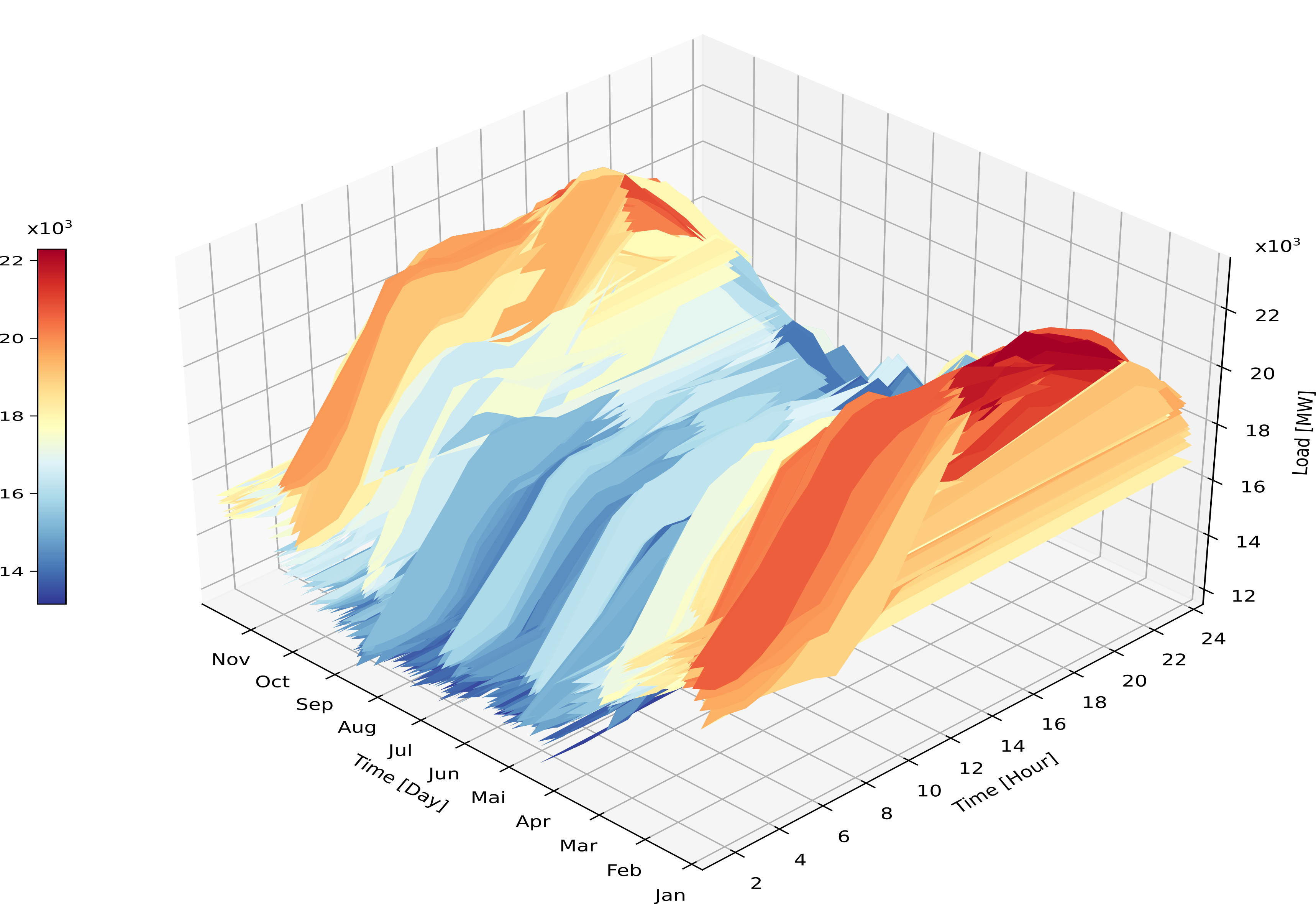}
    \caption{Electricity demand data of Ukraine from January 01.2019 to December 31.2019.}
    \label{fig:weeklyMA}
\end{figure}

The medium and short-term models characterize the seasonal pattern of the time series. Figure \ref{fig:weeklyMA} displays the electricity demand in Ukraine in 2019 based on an an hourly resolution for each day of the year. The mid-term seasonal load pattern - with an increase of demand in winter - can be perceived by the red areas in daily resolution. The recurring transversal of the daily short-term peak load is shown by the x-axis in hourly resolution.

\begin{figure}[ht]
    \centering
    \includegraphics[width=0.8\textwidth]{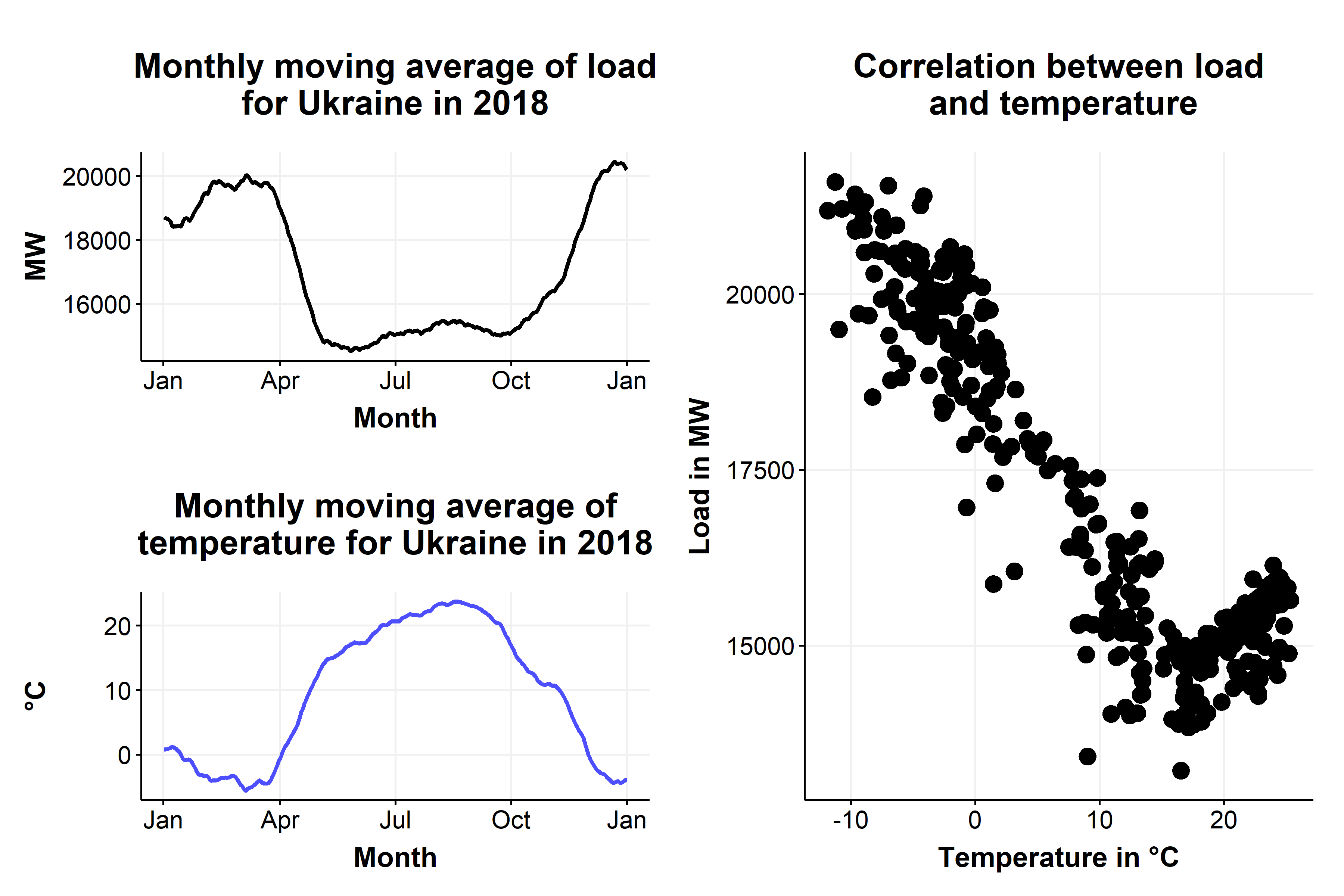}
    \caption{Comparison of Temperature and Load.}
    \label{fig:temperatureMA}
\end{figure}

A prominent variable to characterize the seasonality of electricity demand is temperature. For this study temperature data is retrieved from \cite{Merra2}. Figure \ref{fig:temperatureMA} displays the monthly moving average of temperature and load in 2018 in the left plot, and a scatter-plot to show the correlation structure between the two variables. The graphs on the left demonstrate that temperature and load follow a non-linear reverse pattern throughout most of the year. Meanwhile, the figure on the right displays a scatter-plot following the shape of a hockey stick, with the highest load levels attained at cold temperatures, and loads for high temperatures slightly above those for milder days. In Ukraine, $5.3 \%$ of residential load results from heating appliances, and only $1.9 \%$ from cooling devices \cite{Urkstat2019}. 

\section{Mathematical Model}
As macroeconomic and demographic variables effect the yearly trend, temperature, the intra-year patterns and calendar variables effect the daily profile. Hence, the model is split into three parts. Consequently, it consists of long, medium and short-term parts. By splitting load into its components, the magnitude of the multiple seasonalities and influence of exogenous variables does not increase over time. Thus, an additive approach is chosen:  
\begin{equation}
\label{eq:complete_model}     
\bar{D}(t(y,d,h)) = \bar{D}_{L}(y)+ \bar{D}_{M}(d,y)+\bar{D}_{S}(y,d,h)
\end{equation}

\noindent where $\bar{D}(t)$ is the hourly electricity demand at time $t(y,d,h)$, and $\bar{D}_{L}(\cdot), \bar{D}_{M}(\cdot)$ and $\bar{D}_{S}(\cdot)$ are the long-, medium-, and short-term components, with a yearly (y), daily (d) and hourly (h) resolution, respectively. 

\subsection{The Long-term Classical Regression Model}
\begin{figure}[ht]
\label{fig:LongTerm}
\centering
\begin{subfigure}{.5\textwidth}
  \centering
  \includegraphics[width=1\linewidth]{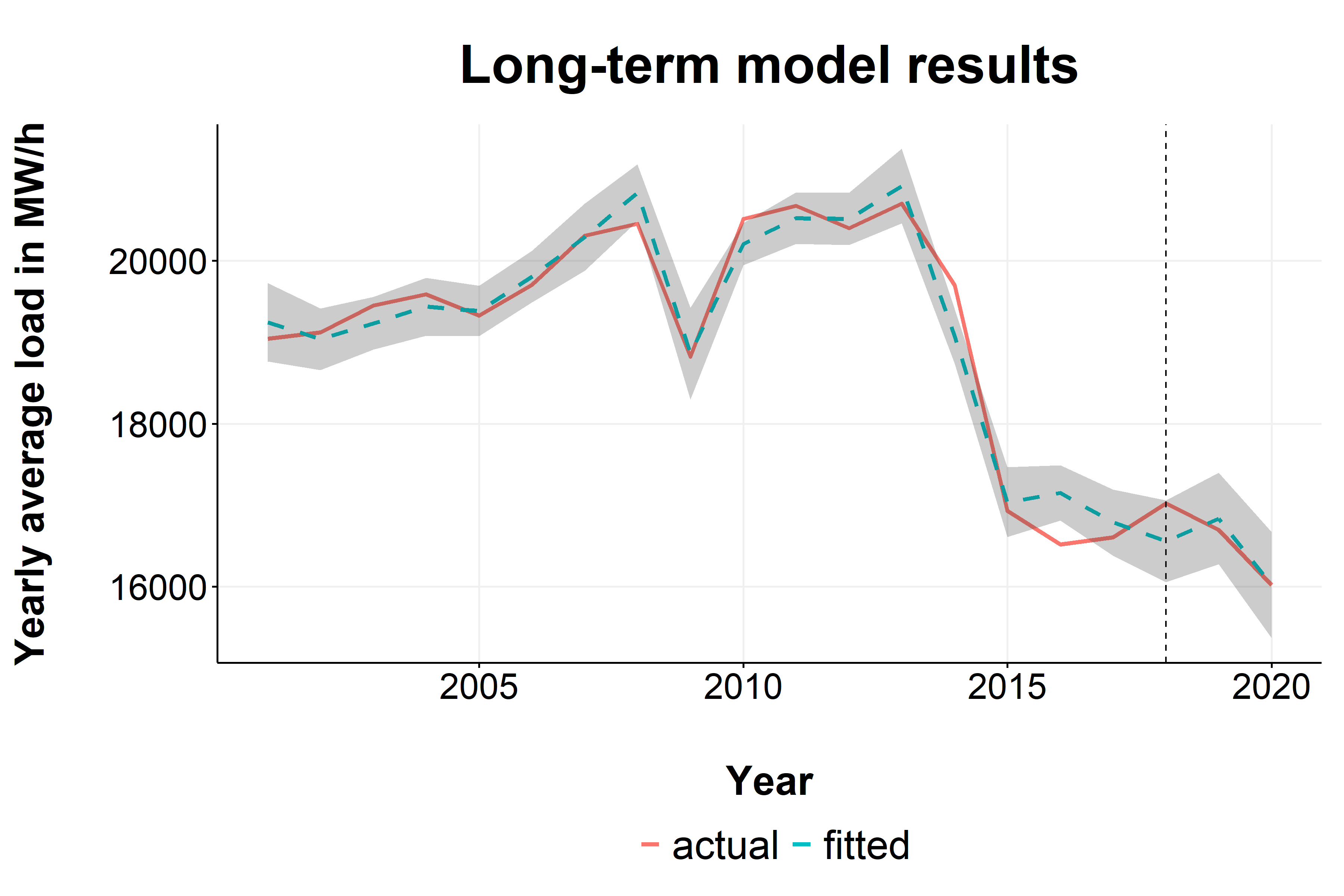}
  \caption{Long-term actual and predicted values}
  \label{thisSubfigure}
\end{subfigure}%
\begin{subfigure}{.5\textwidth}
  \centering
  \includegraphics[width=1\linewidth]{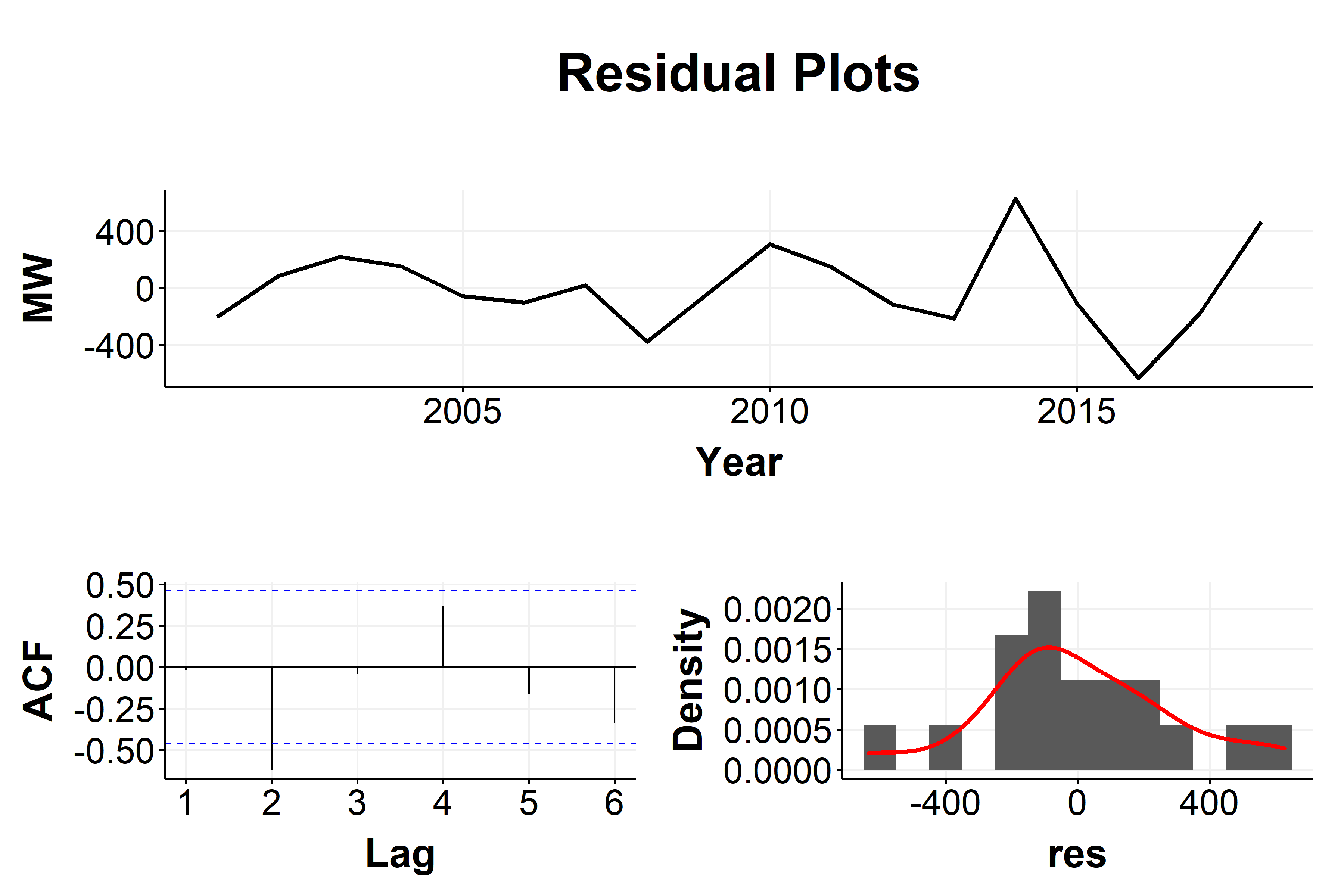}
  \caption{Residuals for the long-term model}
  \label{fig:LongTermResiduals}
\end{subfigure}
\caption{(a) Comparison of long-term model predictions (dashed line) and actual values. The vertical, dashed line marks the transition from training to test set. The grey area indicates the 95 \% confidence interval. (b) The upper residual plot shows the absolute residual values for each year in the training set, the bottom left plot shows the auto-correlation function of the residuals and the bottom right plot shows the residual density distribution.    }
\label{fig:test}
\end{figure}

We first created a long-term model, which predicts averages of hourly load per year based on macroeconomic factors. We denote the average of hourly load for year $y\in\{2001, \dots, 2018\}$ by $\bar{D}_{L}(y)$. If we let $t_1\in\{1, 2, \dots, 8\}$ it is defined by: 
$$\bar{D}_{L}(t_1)=\frac{1}{8760}\sum_{j=1+8760(t_1-1)}^{8760*t_1} D_h(j)$$
where $D_h$ is the given hourly demand in the data set. To prepare the data set we calculated hourly averages for the years 2013 until 2020 from the data obtained from \cite{Ukrenergo2020a}. Since eight years are not sufficient to evaluate the long-term relationship, hourly averages for the years 2001 and 2012 were obtained from \cite{Global}. Additionally, macroeconomic factors for Ukraine for the years from 2000 to 2020 were acquired from \cite{Worldbank}.
To select the most appropriate long-term model, a total of 18 independent series of macroeconomic factors ranging from 2001 to 2018 were taken into account and multiple linear regression models (LM) were calculated for all possible combinations, as in \cite{Amarawickrama2008, Pielow2012}. The data for 2019 and 2020 were not included, to use them for the out-of-sample fit check. For 18 predictor values, this leads to $2^{18}$ linear models. These models were ranked by AICc and the best 1000 models were subjected to a k-fold (k=5) cross validation to avoid over-fitting. The lowest RMSE and MAE values of the cross validation results were taken as baseline and all models, which exceeded the baseline values by more than 150 \% , were discarded.
The remaining models were then used to predict the test set of 2019-2020. The models with the lowest maximum distance between all of the actual and predicted values were chosen and checked for normality of residuals and multicollinearity. The selected long-term model is defined as:

\begin{equation} \label{eq:LongTerm_actual}
\begin{split}
\bar{D}_L(y) = \beta_{L,1}+ \beta_{L,2}GDP_{\delta}(y)+\beta_{L,3}IND(y) 
+\beta_{L,4}GRO(y)\\ +\beta_{L,5}SERV(y)+w(y)
\end{split}
\end{equation}

where $\beta_{L,i}$ are regression coefficients with $i\in\{1, \dots, 5\}$, $GDP_{\delta}$ is the annual GDP deflator in \% , $IND$ the industrial value added to GDP in constant 2015 USD , $GRO$ the annual GDP growth rate in \%, and $SERV$ the added service value in current USD. Lastly, $w(y)$ is a noise process with zero mean and variance $\sigma^2_w$.\\
The predicted and actual yearly averages of hourly load are shown in Figure \ref{thisSubfigure} and the residual structure is illustrated in Figure \ref{fig:LongTermResiduals}. To complement the graphical methods, we performed a Shapiro-Wilk normality test and with a p-value of 0.62 the error process follows a normal distribution with $\sigma_w^2=341.1$.

\subsection{Medium-term Model}
Let $D_d(t_2, t_1)$ denote the hourly average of day $t_2\in\{1, 2, \dots, 365\}$ of year $t_1$, which is defined as:
$$\bar{D}_{d}(t_2,t_1)=\frac{1}{24}\sum_{j=1+24(t_2-1)+8760(t_1-1)}^{24t_2+8760(t_1-1)} D_h(j)$$
where $D_h$ is the given hourly demand in the data set. 
Medium-term demand is defined as the difference between daily average of hourly load and the yearly average of hourly load:

\begin{equation} \label{eq:MidTerm_demand}
\bar{D}_M(t_2, t_1)= \bar{D}_{d}(t_2,t_1)-\bar{D}_{L}(t_1).
\end{equation}
By subtracting the hourly yearly average, the time series is corrected for the long-term trend. As a result, $\bar{D}_M$ has a yearly mean of zero. 
As in \citet{Sotiropoulos2013}, we split our mid-term model into a linear regression and a stochastic component :
\begin{equation} \label{eq:MidTerm_general}
\bar{D}_M(t_2, t_1)= d_{M}(t_2,t_1)+\xi_{M}(t_2, t_1)
\end{equation}
The linear regressive part $d_M(t_2, t_1)$ refers to the trend and seasonal component, while the stochastic model $\xi_M(t_2, t_1)$ is fitted to the remainder. Here, it is assumed that the residuals of the linear regressive part still hold relevant information and an underlying structure can still be captured by ARIMA or ML methods.

In the medium-term model, with daily granularity, electricity demand is mainly influenced by temperature, season and day of the week. Due to the non-linear relationship between load and temperature, daily average temperature values were transformed to heating and cooling days as introduced by \cite{train1984billing} and adopted in \cite{PARDO2002,Pielow2012}. Heating days are defined as $HD(t_2,t_1)=max\{T_{ref}-T(t_2,t_1),0\}$
and cooling days as $CD(t_2,t_1)=max\{T(t_2,t_1)-T_{ref},0\}$. With $T$ being temperature. The value for $T_{ref}=18.33$ is taken from \cite{Pielow2012}. Terms of HD and CD values up to the third polynomial were included. Lagged HD, CD and temperature values with a lag of one day and two days were added to the predictor variables, to account for consecutive days of hot or cold weather. To capture the seasonal load pattern and avoid perfect collinearity, dummy variables for the first eleven months $(ToM)$ are constructed. Likewise, to obtain the weekly seasonality, we consider a dummy variable for each of the first six weekdays $(ToD)$ and another dummy variable for holidays ($H$). A total of 32 predictor variables are taken into consideration for the linear regressive mid-term model. This leads to $2^{32}$ possible linear model combinations. To lower the computational effort, when selecting the best model, we excluded the dummy variables for each weekday (6 in total) and calculated the best linear model of the remaining $2^{26}$ possible combinations with regard to the smallest AICc. Then excluded predictors are added stepwise to obtain the model that fits best. The resulting linear regression model is defined by:

\begin{equation} \label{eq:MidTerm_det}
\begin{split}
\bar{d}_M(t_2, & t_1)  =  \beta_{M,1}+ \beta_{M,2}T(t_2, t_1)+\beta_{M,3}T_{i-2}(t_2, t_1)+\beta_{M,4}CD^2(t_2, t_1) \\
& +\beta_{M,5}CD^3(t_2, t_1) +\beta_{M,6}CD_{i-2}(t_2, t_1)   +\beta_{M,7}HD(t_2, t_1) \\ & +\beta_{M,8}HD^2(t_2, t_1) +\beta_{M,9}HD^3(t_2, t_1) 
+\beta_{M,10}HD_{i-2}(t_2, t_1)  \\
& +\beta_{M,11}H(t_2, t_1)  +\sum_{l=1}^{6}\beta_{M,l}~ToD_{l}(t_2, t_1)\\
&+\sum_{k\in\{3,4,6,8,10\}}\beta_{M,k}~ToM_{k}(t_2, t_1)
\end{split}
\end{equation}
Where $\beta_{M,j}$ are regression coefficients for the intercept and temperature related regressors for $j\in\{1, \dots, 11\}$;  $\beta_{M,l}$ are regression coefficients for each type of day for $l\in\{1, \dots, 6\}$; and $\beta_{M,k}$ are five regression coefficients for the respective months. 

\begin{figure}[ht]

\centering
\begin{subfigure}{.5\textwidth}
  \centering
  \includegraphics[width=0.9\linewidth]{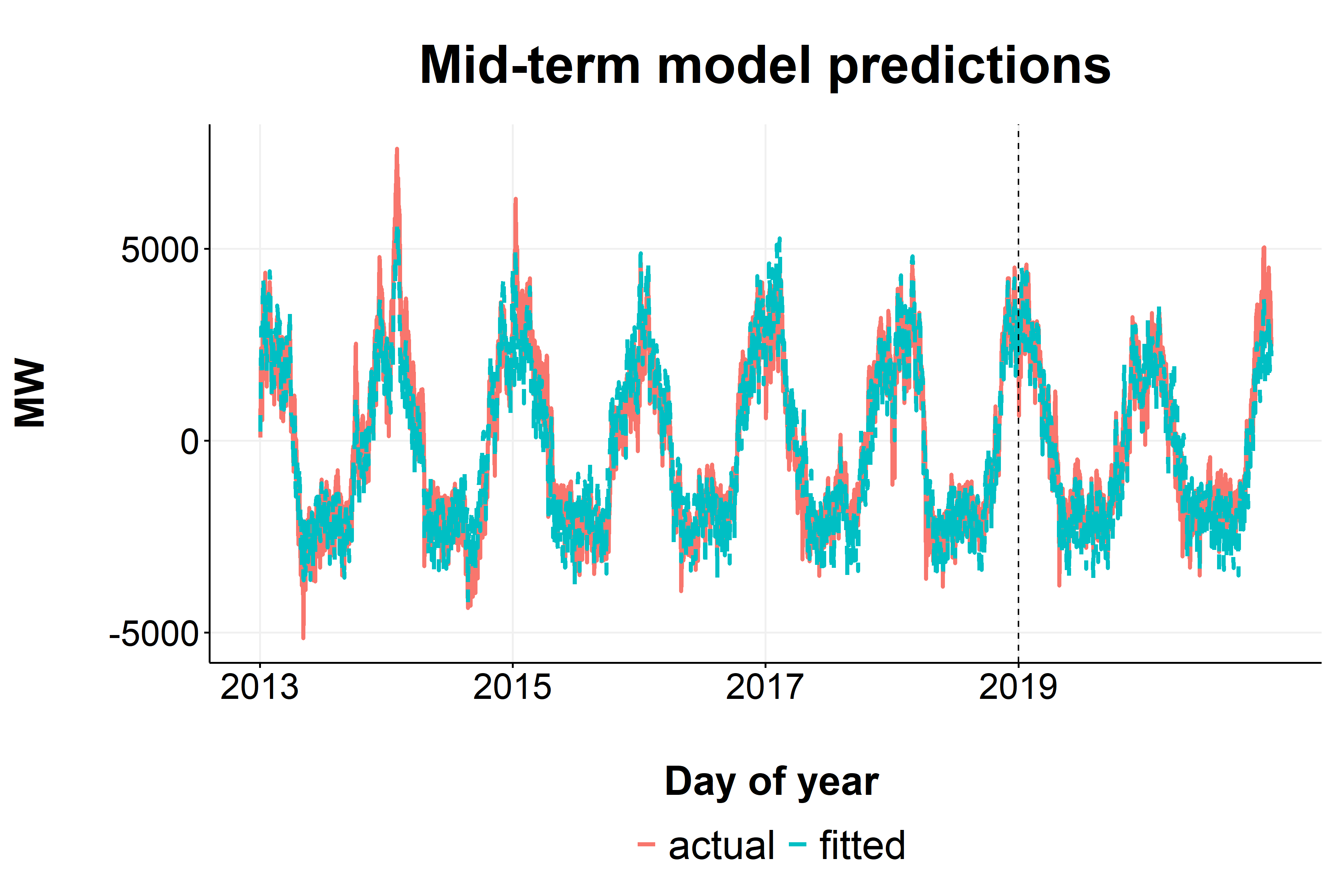}
  \caption{Linear regressive mid-term model predicted and actual values.}
  \label{fig:MidTerm}
\end{subfigure}%
\begin{subfigure}{.5\textwidth}
  \centering
  \includegraphics[width=0.9\linewidth]{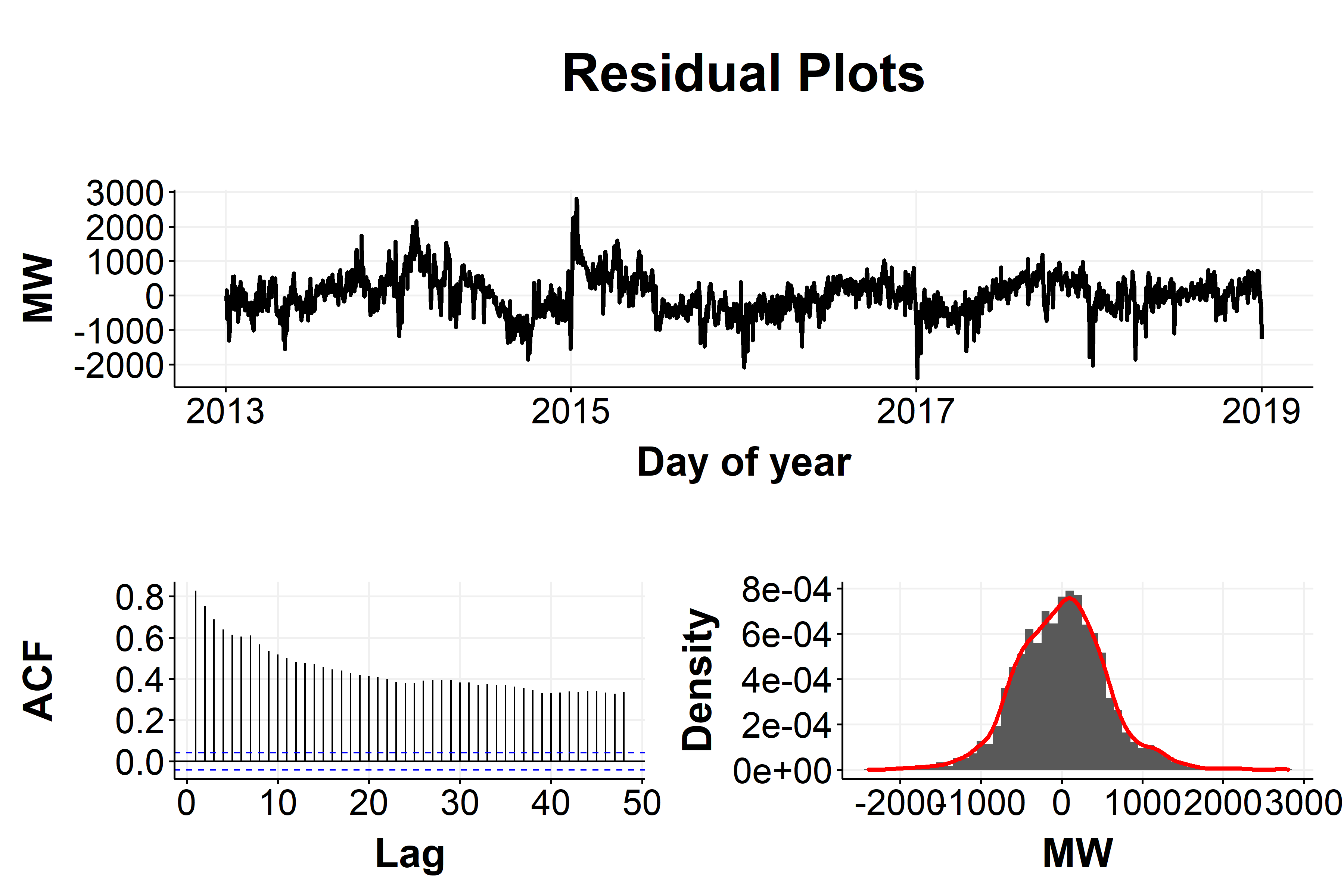}
  \caption{Residual analysis of the linear regressive mid-term model.}
  \label{fig:MTermResiduals}
\end{subfigure}
\caption{(a) Comparison of mid-term model predictions (dashed blue line) and actual values. The dashed, vertical line marks the transition from training to test set.  (b) The upper residual plot shows the residual values for each day in the training set, the bottom left plot shows the auto-correlation function of the residuals and the bottom right plot the residual density distribution.}
\label{fig:MidTerm_det_combined}
\end{figure}

Figure \ref{fig:MidTerm_det_combined} shows the predicted and actual values for the mid-term model, defined in Equation \ref{eq:MidTerm_det}. The residuals are approximately normally distributed with a zero mean and a standard deviation of 590 MW. However, the auto-correlation function (ACF) plot in Figure \ref{fig:MTermResiduals} displays significant auto-correlation. Therefore a stochastic ARIMA model and a ML approach are used to predict the remaining underlying structure of the residuals that differs from white noise. 


\subsubsection{Medium-term stochastic AR component}\label{sec:Medium_stochastic}

The AR part of the medium-term model is fitted to the residuals of the linear regressive component. An augmented Dickey-Fuller (ADF) test as well as a Kwiatkowski-Phillips-Schmidt-Shin (KPSS) test  were performed, which revealed that the series of residuals is non-stationary. Consequently, we difference to achieve stationarity and an ARIMA model approach is taken. 
A typical ARIMA(p,d,q) model can be expressed by: 
\begin{equation}\label{arima}
    \varphi_p (B)(1-B)^d z_t=\theta_q(B) \varepsilon_t  
\end{equation}
where $z_t$ represents a nonstationary time series at time $t$, $ \varepsilon_t$ is a white noise, $d$ is the order of differencing, $B$ is the backward shift operator defined by $Bz_t=z_{t-1}$ and $\varphi_p(B)$ is the autoregressive operator defined as:
\begin{equation}
    \varphi_p(B) = 1- \varphi_1B - \varphi_2 B^2- \dots - \varphi_p B^p.
\end{equation}
Also, $\theta_q(B)$ is the moving average operator defined as:
\begin{equation}
    \theta_q(B) = 1- \theta_1B - \theta_2 B^2- \dots - \theta_q B^q
\end{equation}

\begin{figure}[ht]
\centering
\begin{subfigure}{.5\textwidth}
  \centering
  \includegraphics[width=0.9\linewidth]{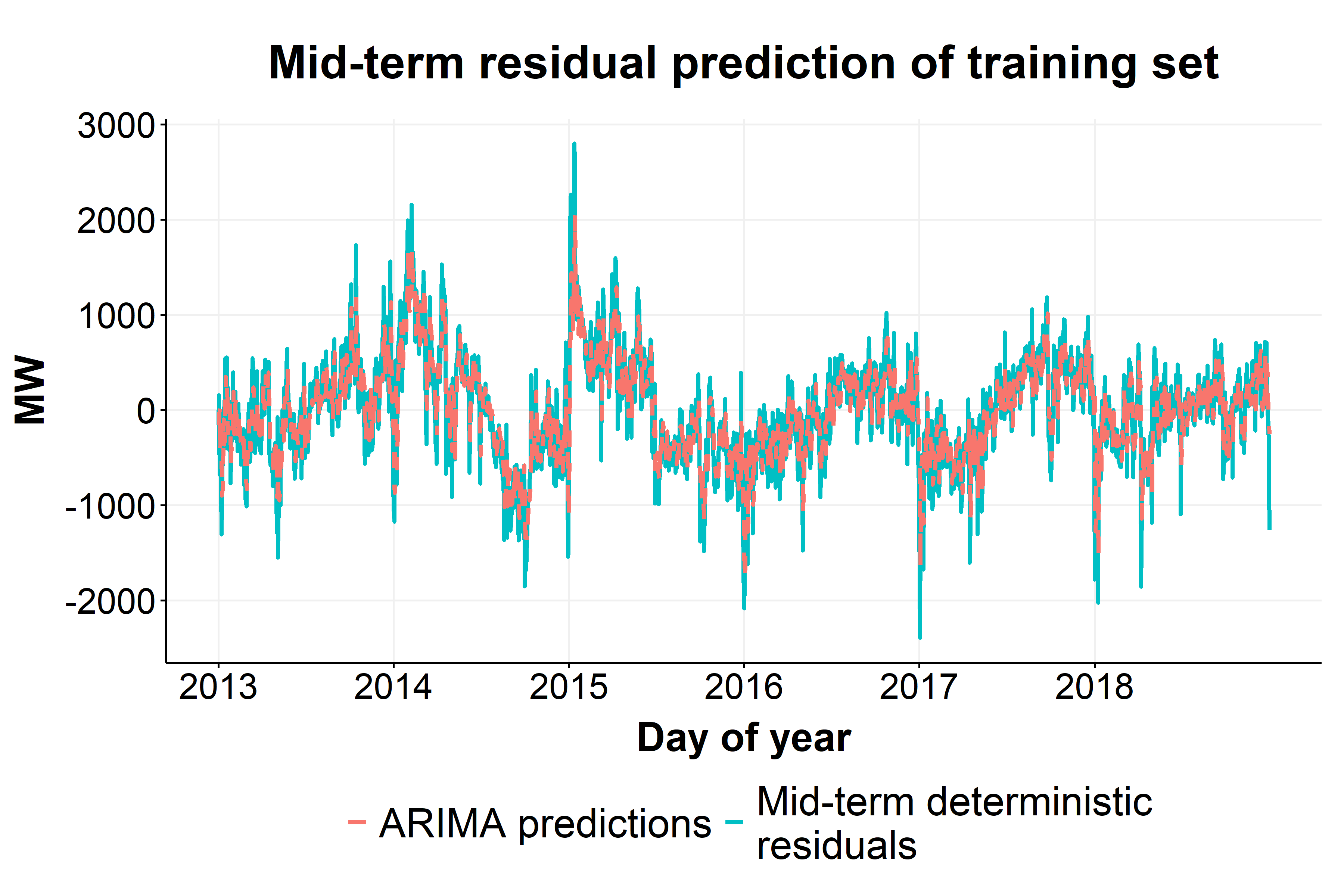}
  \caption{Mid-term deterministic residuals and predicted residual values by ARIMA modeling.}
  \label{fig:Arima_values_training}
\end{subfigure}%
\begin{subfigure}{.5\textwidth}
  \centering
  \includegraphics[width=0.9\linewidth]{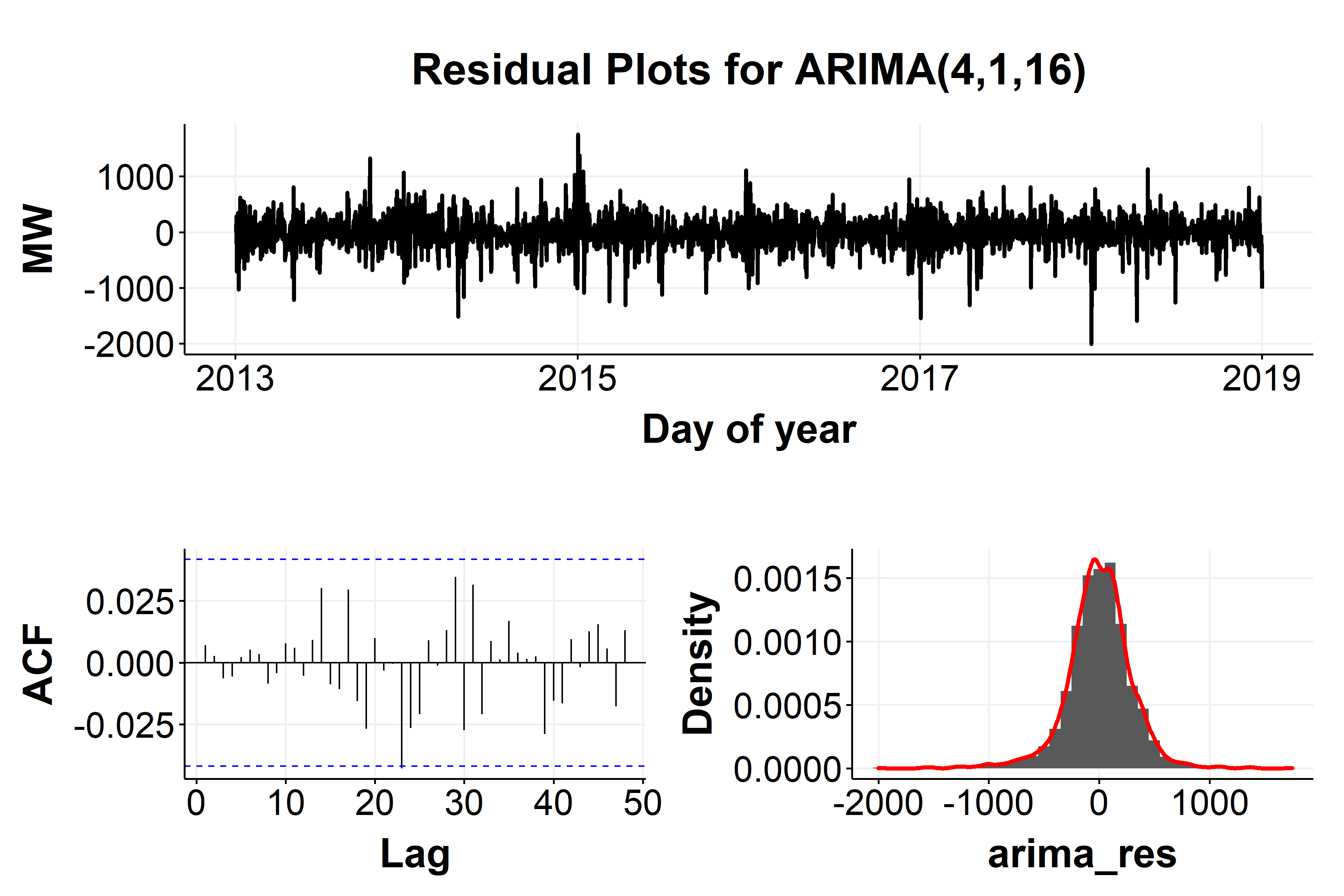}
  \caption{Residuals of ARIMA model predictions}
  \label{fig:Midterm_stoch_residuals}
\end{subfigure}
\caption{In-sample ARIMA prediction and residual structure.}
\label{fig:Arima_training_set}
\end{figure}

To determine an appropriate model, the following method with respect to replicability for other datasets and custom automatization possibilities was applied. Firstly, the lowest order of differencing needed for the series to become stationary was identified. This is done by utilizing the \textit{auto.arima} function within the \textit{R forecast} package \citep{forecast1, forecast2}. The differenced time-series is then tested with an ACF and a partial auto-correlation function (PACF) to determine the highest orders of p and q respectively for which significant auto-correlation is left. The obtained values are then used as an upper limit for a step-wise grid search with varying p and q values starting from 1 respectively. The upper boundaries for p and q values are increased by 2 to allow for an inspection of models with slightly higher complexity than the orders determined by the ACF and PACF.
The d value given by the \textit{auto.arima function} is kept the same as it is already the optimal order of differencing needed for stationarity.\\
ARIMA models for all assigned p, d and q values are then calculated using conditional-sum-of-squares estimation to find starting values, and afterwards maximum likelihood estimation for the respective
AR and MA coefficients.
\\
Most ARIMA models show very high prediction accuracy in the training set from 2013 to 2018 but significantly worse prediction accuracy for the test set of 2019 and 2020. Therefore, the grid search is set to choose the ARIMA model with the best prediction accuracy for the test set, rather than the model with the lowest AICc for the training set. \\
After the best ARIMA model is determined, a verification is conducted to determine if forecast accuracy can be improved further by adding exogenous regressor variables. For this purpose within-group uniformity and between-group homogeneity of variances are tested with multiple Kolmogorow-Smirnow-tests (KS) and a Levene-test respectively, for each predictor variable used in the linear regressive mid-term model. Significant deviations from within-group uniformity in the mid-term residuals are found for the independent variables June, August and October. Therefore those 3 predictor variables were included as external regressors in the ARIMA model. \\
With the above discussed method an ARIMA(30,1,25) with external regressors was determined to be the best fitting model to describe the residuals left by the mid-term deterministic model. Figure \ref{fig:Arima_training_set} displays the residual prediction of the chosen ARIMA model for the training set of 2013-2018 as well as the respective residuals. Judging from Figure \ref{fig:Arima_values_training}, the residual time series can be described very well by the chosen ARIMA, while the ACF plot in Figure \ref{fig:Midterm_stoch_residuals} shows that there is no significant auto-correlation left between the residuals.

\begin{figure}[ht]
    \centering
    \includegraphics[width=0.7\textwidth]{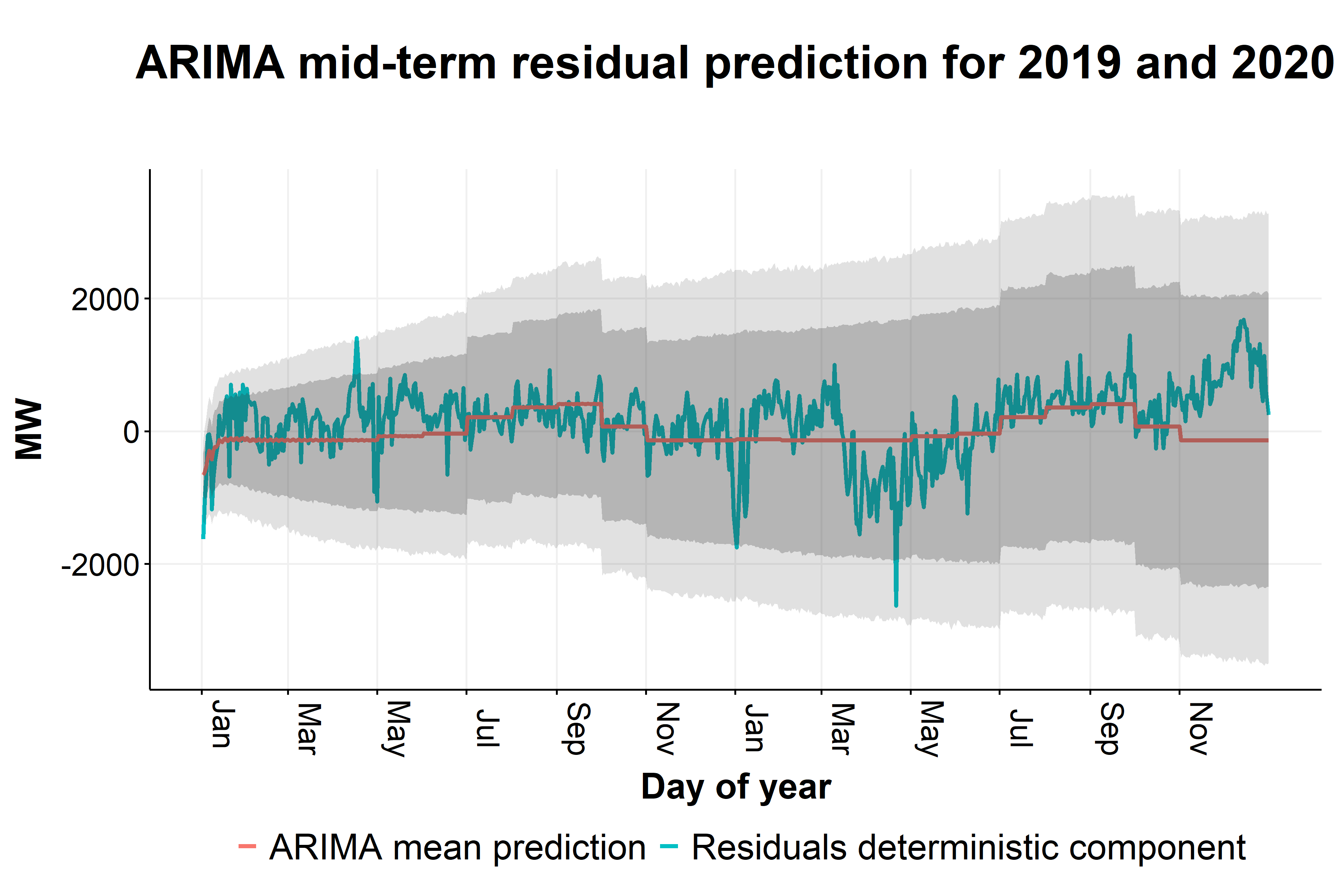}
    \caption{Out-of sample ARIMA mid-term residual forecast for 2019 and 2020. The dark grey area shows the 80 \% confidence interval and the light grey area the 95 \% confidence interval.}
    \label{fig:Arima_forecast}
\end{figure}

The forecast for the test set of 2019 and 2020 is presented in Figure \ref{fig:Arima_forecast}.
It is shown that the residuals of the linear regression $\bar{d_M}$ mid-term model are well within the 80 \% confidence interval of the ARIMA forecast. The only point in time where the values are beneath the 95 \% confidence interval is in April 2020. This is due to a decrease in energy consumption at the early stages of the COVID-19 pandemic. According to Figure \ref{fig:Arima_forecast}, it is clear that the ARIMA model can provide useful information on the magnitude of prediction errors for the mid-term regression model, but lacks the possibility to accurately predict actual values in future time ranges that exceed the lag order of the model. 

In a shorter prediction time horizon, for instance a two-week or monthly forecast, this approach will greatly enhance forecasting accuracy, wherein for longer time horizons, only a small improvement can be expected. This is caused by the fact that after the forecasted time horizon exceeds the lag order of the model, predictions are based on former predicted values without validation. Therefore only the forecasted mean can be safely assumed. To overcome this limitation and test if the residual time series can be approximated further with a non-linear approach, a machine learning LSTM model is trained and tested.

\subsubsection{Medium term LSTM component} \label{sec:LSTM}
Long short-term memory (LSTM) is a type of Recurrent Neural Network (RNN) developed by \cite{LSTM}. These types of neural networks have a special feedback connection that can store information of recent inputs. Basic RNN have the problem of vanishing gradients on long-term time-dependencies. To solve this issue, LSTM were introduced by Schmidhuber and Hochreiter in 1997 \cite{LSTM}.
The general LSTM architecture consists of a memory cell in which information can be stored and is connected to a neuron by three different gates: the forget gate, input gate and output gate. The forget gate controls which information is stored in the cell from the last state ($c_{t-1}$) according to Equation \ref{eq:forget gate}. The information used for the current state is determined by the input gate (Equation \ref{eq:input gate}) and the output gate controls the information necessary for the output (Equation \ref{eq:output gate}):
\begin{equation} \label{eq:forget gate}
f_t = \sigma(W_f * [h_{t-1}, X_t] + b_f)
\end{equation}
\begin{equation} \label{eq:input gate}
i_t = \sigma(W_i * [h_{t-1}, X_t] + b_i)
\end{equation}
\begin{equation} \label{eq:output gate}
o_t = \sigma(W_o * [h_{t-1}, X_t] + b_o)
\end{equation}

Where $W_x$ and $B_x$ denote the different weights and biases of the neuron, $\sigma$ is a sigmoid activation function, $h_{t-1}$ is the hidden layer output at time step $t-1$ and $X_t$ is the input vector at each time step.
The current hidden state $c_t$ is described by:

\begin{equation} \label{eq:current state part}
\bar{c}_t = tanh(W_c * [h_{t-1}, X_t] + b_c)
\end{equation}
\begin{equation} \label{eq:current state}
c_t = ft*c_{t-1} + i_t * \bar{c}_t 
\end{equation}

where $W_c$ and $b_c$ denote the weights and biases of the current gate with the hyperbolic tangent activation function denoted by $tanh$. The output of the LSTM layer is then:
\begin{equation} \label{lstm output}
h_t = o_t * tanh(c_t)
\end{equation}

\begin{figure}[H]
    \centering
    \includegraphics[width=0.7\textwidth]{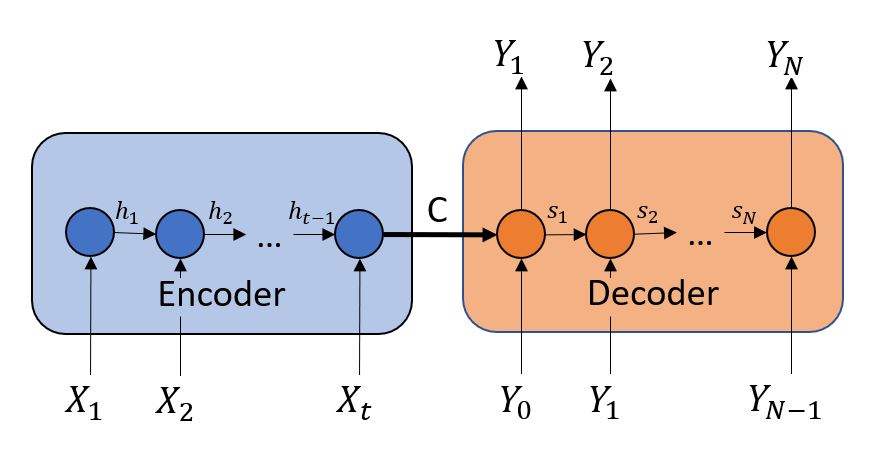}
    \caption{Structure of a Sequence-to-Sequence model}
    \label{fig:Seq2Seq-Model}
\end{figure}

For this study, a Sequence-to-Sequence (S2S or Seq2Seq) algorithm is used as input vector for the LSTM model. Seq2Seq models consist of two parts: an encoder and a decoder (Figure \ref{fig:Seq2Seq-Model}). 

For an input sequence $X = [X_1, X_2, ..., X_t]^T$, the LSTMs used in this study encode a representation of the data within the hidden states of the layer $h = [h_1, h_2, ..., h_t]^T$. This is done by calculating a hidden state $h_t$  by using the previous hidden state of the neuron $h_{t-1}$, current input $X_t$ and the layers recurrent computations $f(\cdot)$ with:

\begin{equation} \label{eq:hidden state calculation}
h_t = f(h_{t-1},X_t)
\end{equation}

The context vector $C$, which is the hidden state of the last time step, is then fed to the decoder to predict future values with:

\begin{equation} \label{eq:decoder prediction}
y_t  = g(s_{t-1},C)
\end{equation}

where $s$ represents the state of the hidden layer, $g(\cdot)$ the layers activation function and $y_t$ the prediction value. 

Longer input sequences to the encoder can lead to difficulties for the decoder in identifying valuable information from the context vector. This is due to the fact that the representation of the data provided resembles a single vector. Attention mechanisms let the decoder focus on the most important parts flexibly. The mechanism used in this study is the Luong Attention mechanism developed by \cite{luong-etal-2015-effective}. For that, an alignment vector $a_t$ is computed by comparing the hidden states from decoder $h_t$ with all hidden states of encoder $h_s$ with:

\begin{equation} \label{eq:allignment vector}
a_{ti} = align(h_t,s_t) = \frac{exp(S(h_t, h_s))}{\sum_{k}S(h_t, h_{sk})}
\end{equation}

For the scoring function $S(h_t, h_{sk})$ Luong et al. propose three different functions:

\begin{equation} \label{eq:scoring functions}
S(h_t, h_{sk})
\left\{\begin{matrix}h_t^Th_s & dot\\h_t^TW_ah_s & general\\v_a^Ttanh(Wa[h_t;h_s]) & concat\\\end{matrix}\right.
\end{equation}
with $v$ as a trainable parameter, which can be learned within the neural network training stage. In this study the tensorflow/keras implementation \cite{tensorflow2015-whitepaper} of attention is used.

\begin{table}[H]
\ra{1.3}
  \begin{center}
    \caption{optimized Seq2Seq Model Parameter}
    \label{tab:LSTM_Model_params}
    \begin{tabular}{c| c c} 
      \toprule 
       \textbf{Parameter Setting}&\textbf{Layer 1} & \textbf{Layer 2}\\
      \midrule

Layertype & LSTM & LSTM \\
Nodes & 118 &  82 \\
Activation & Sigmoid &  Relu \\
Dropout & 0.385 &  0.08 \\
Output Size &  1 \\
\hline
Optimizer & \multicolumn{2}{c}{ADAM} \\
Epochs & \multicolumn{2}{c}{300} \\
Batch Size & \multicolumn{2}{c}{64}\\ 
Learning Rate & \multicolumn{2}{c}{0.00074}\\
Input Length & \multicolumn{2}{c}{584}\\
Output Length & \multicolumn{2}{c}{730}\\
      \bottomrule
    \end{tabular}
  \end{center}
\end{table}

\begin{figure}[ht]
    \centering
    \includegraphics[width=0.7\textwidth]{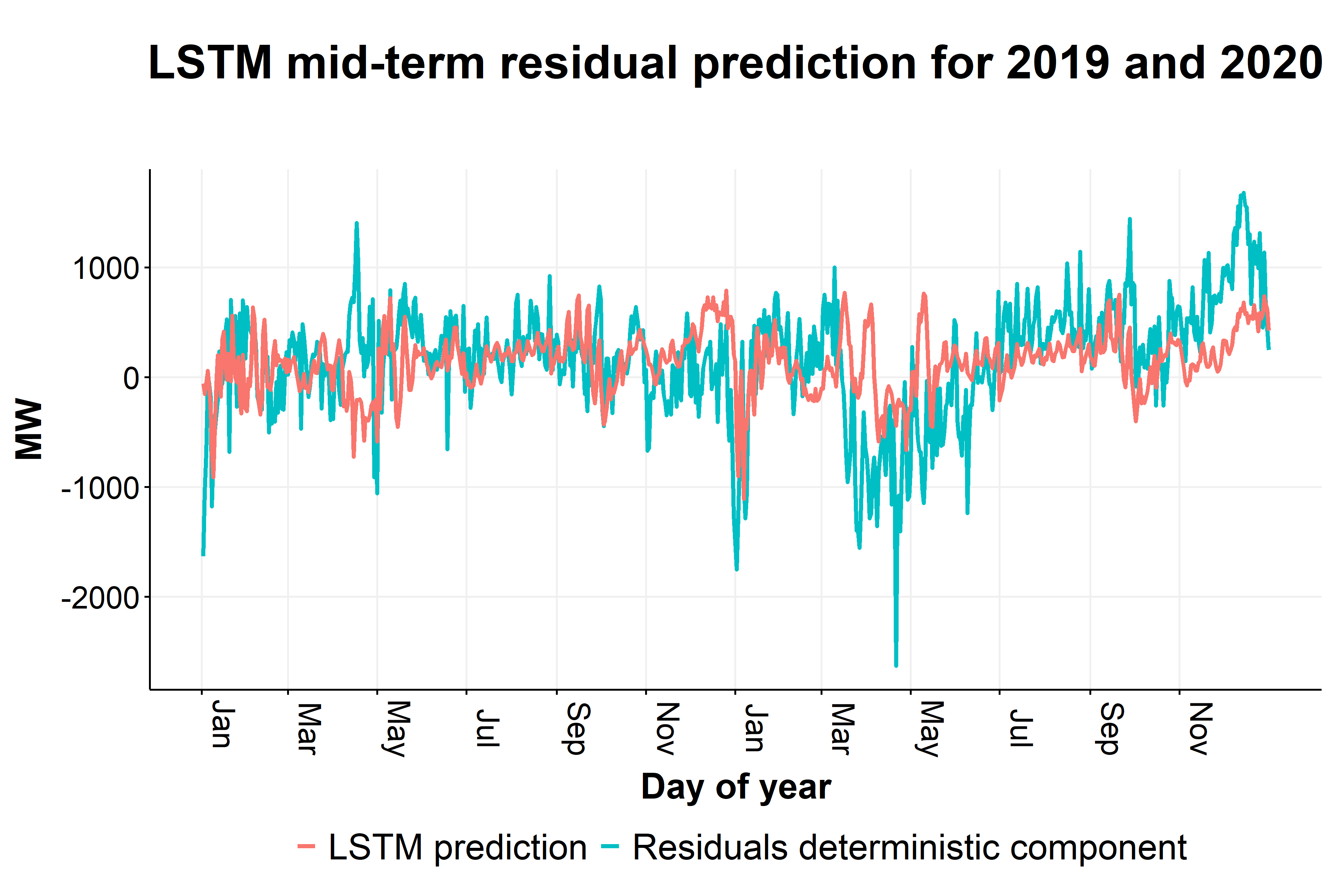}
    \caption{Residuals of the linear regression mid-term model and out-of sample LSTM mid-term residual forecast for 2019 and 2020.}
    \label{fig:LSTM_forecast}
\end{figure}

The results of the LSTM model approach are illustrated in Figure \ref{fig:LSTM_forecast}. Compared to the ARIMA forecast shown in Figure \ref{fig:Arima_forecast}, a much more dynamic residual time series is forecast, capturing parts of the residual structure. Adding the forecasted residuals to the predictions of the linear regression mid-term model decreases the residual sum by about 13.2 \%.

\subsubsection{Hybrid mid-term model}
Three different hybrid models were tested and evaluated. A linear regressive model (LM) with added ARIMA residual prediction, a LM with added LSTM residual prediction, and a LM with a combination of added LSTM and ARIMA residual prediction. Selected error metrics and the effect on the overall residual sum for each approach can be found in Table \ref{tab:midterm_errors}. For the residual sum, we calculated the sum of all residuals for all timesteps and took the value obtained from the LM forecast as baseline (100 \%). We then evaluated if the overall residual sum for the respective hybrid forecasts is smaller (below 100 \%) or bigger. The LM with added LSTM forecast yielded the best results. The complete mid-term model defined in Equation \ref{eq:MidTerm_general} is therefore the sum of $d_M(t_2,t_1)$ defined in Equation \ref{eq:MidTerm_det} and the LSTM RNN model defined in section \ref{sec:LSTM}.

\begin{table}[H]

\ra{1.3}
  \begin{center}
    
    \caption{Comparison of root-mean-square error (RMSE), mean-average error (MAE), mean-average-scaled error (MASE) and effect on the residual sum for the midterm load prediction for the test set of 2019 and 2020.}
    \label{tab:midterm_errors}
    \begin{tabular}{@{}l|cccc@{}} 
      \toprule 
       \textbf{}&\textbf{RMSE}  & \textbf{MAE}  & \textbf{MASE} & \textbf{Residual Sum} \\
      \textbf{}&\footnotesize{[MW]}  & \footnotesize{[MW]}  & \textbf{}
      & \textbf{\%} \\
      \midrule
LM  &  552.8 &  430.8 & 0.495 & 100 \\
LM+ARIMA     &  533.4  & 399.6 & 0.459 & 92.8 \\
LM+LSTM   &        500.6 &  374.0 &   0.429& 86.8 \\
LM+ARIMA+LSTM  &        504.0 &  382.9 & 0.440 &88.9 \\
      \bottomrule
    \end{tabular}
  \end{center}
\end{table}

\begin{figure}[H]
\centering
\begin{subfigure}{.5\textwidth}
  \centering
  \includegraphics[width=0.9\linewidth]{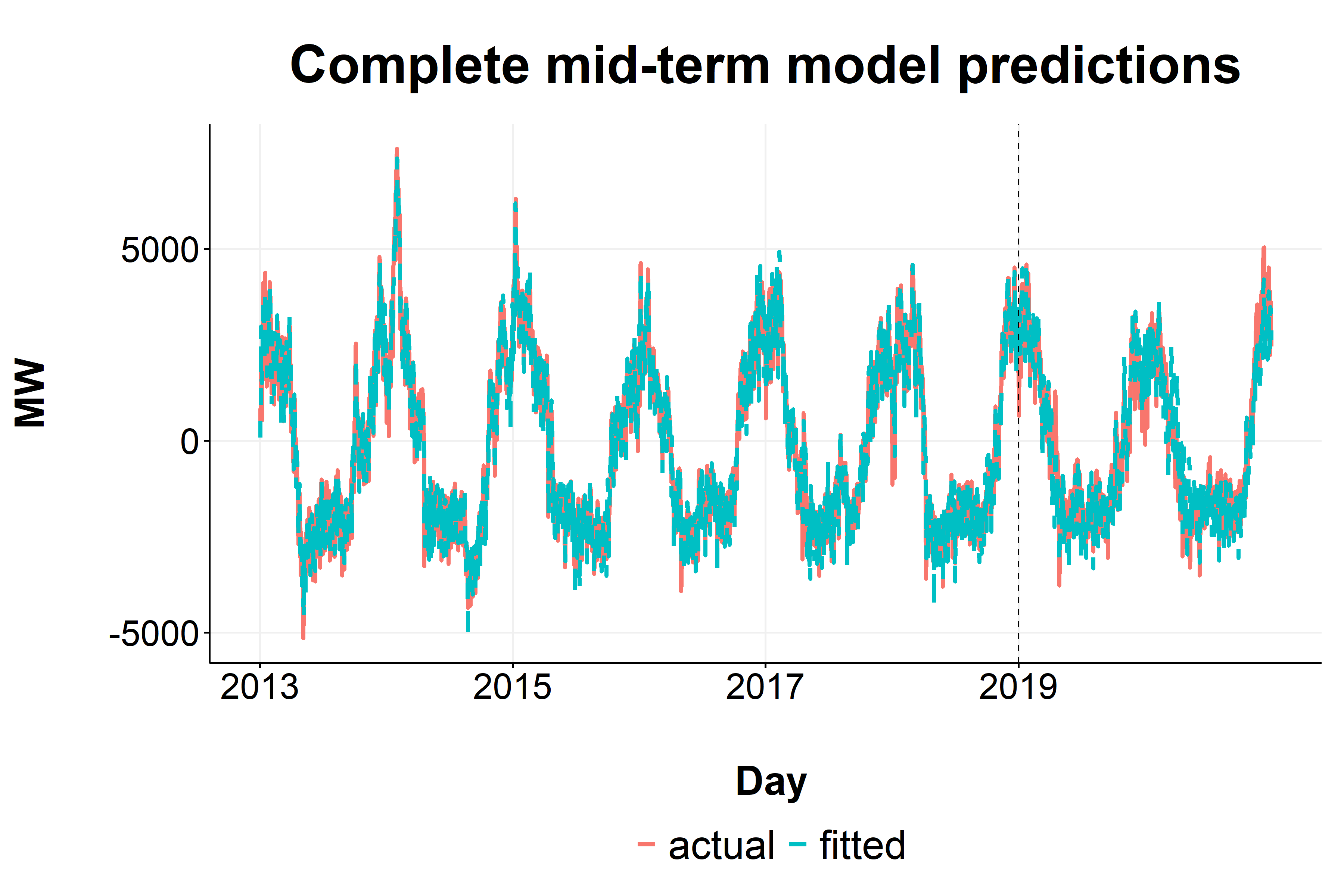}
  \caption{}
  \label{fig:Mid-term_fulla}
\end{subfigure}%
\begin{subfigure}{.5\textwidth}
  \centering
  \includegraphics[width=0.9\linewidth]{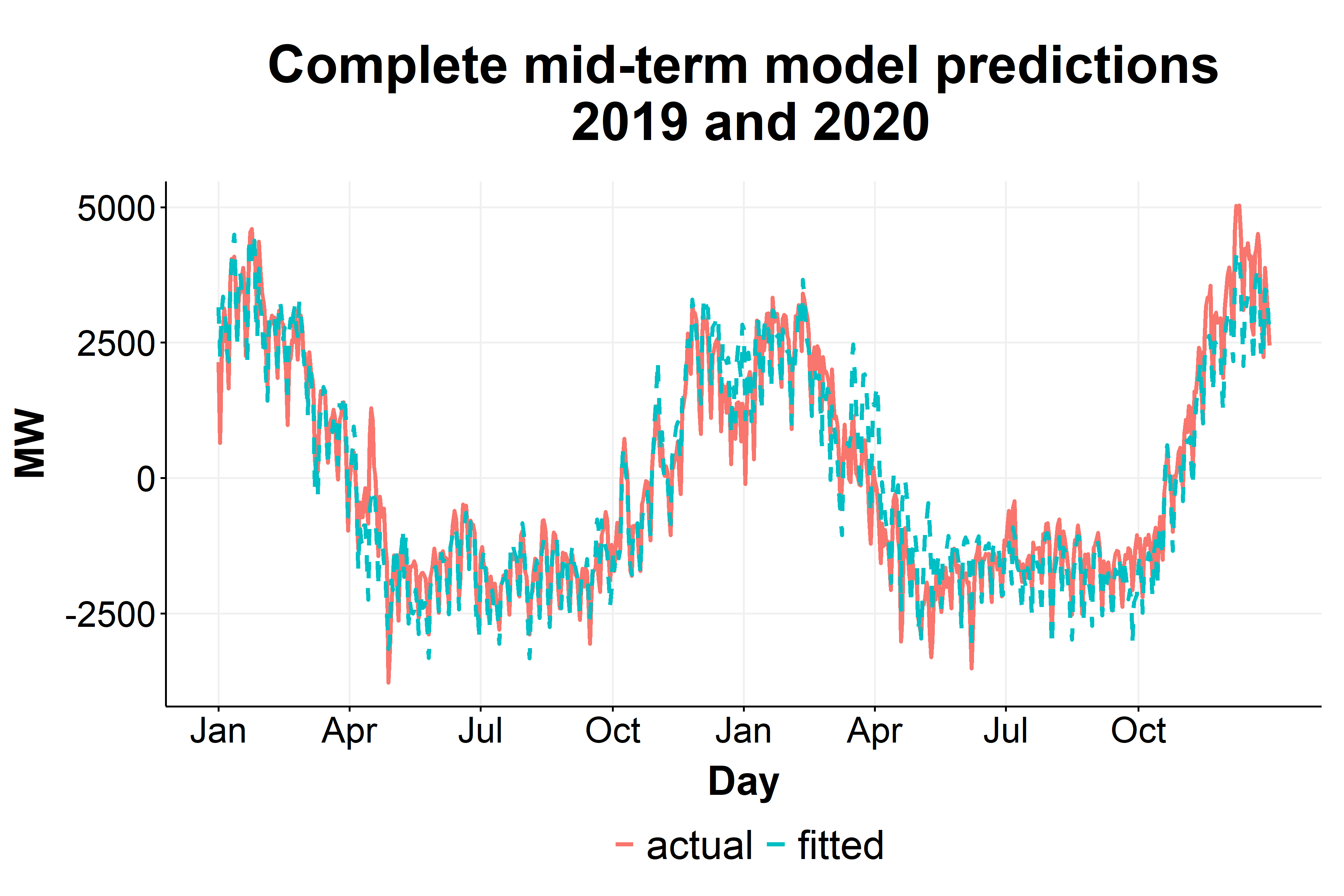}
  \caption{}
  \label{fig:Mid-term_fullb}
\end{subfigure}
\caption{Complete mid-term demand prediction for the whole data set (a) and the test set (b). The vertical dashed line in (a) marks the transition between training and test set.}
\label{fig:Mid-term_fullmodel}
\end{figure}

The complete mid-term forecasting results are shown in Figure \ref{fig:Mid-term_fullmodel} where Figure \ref{fig:Mid-term_fulla} displays the full forecast and \ref{fig:Mid-term_fullb} shows the predictions for the test set of 2019 and 2020 in more detail.

\subsection{Short-term model}

The short-term model is defined as the difference of the yearly and daily averages to hourly load: 
$$\bar{D}_s(t_3,t_2, t_1)=D_h(8760(t_1-1)+24(t_2-1)+t_3)-\bar{D}_M(t_2,t_1)-\bar{D}_L(t_1)$$
where $t_3\in\{1, 2, \dots 24\}$.
By subtracting the yearly and the daily hourly averages from the consumption values, the load pattern specific to the time of day remains. Similar to the mid-term model, the short-term model is defined as the sum of a linear regressive model ($d_{S}(t_3,t_2,t_1)$) and a stochastic part ($\xi_{S}(t_3,t_2, t_1)$):   

\begin{equation} \label{eq:ShortTerm_general}
\bar{D}_S(t_3,t_2, t_1)= d_{S}(t_3,t_2,t_1)+\xi_{S}(t_3,t_2, t_1)
\end{equation}


The daily load profile is mainly influenced by the season, i.e. the month and the type of day. Dummy variables for these characteristics are already included in the medium-term model. As all sub-models are combined into a final additive model, these variables cannot be included additionally in the short-term model. In order to account for these effects, different regression models are fitted for each combination of type of day and month respectively and are added afterwards to predict the full time horizon. The only predictor variable that is included in the linear regressive part, is a dummy variable ToH (0/1) for each hour of the day. Since all the dummy variables are binary only 23 as opposed to 24 hours are included as linear regressor variables. Consequently, the linear regressive short-term model is defined as:

\begin{equation} \label{eq:shortterm_deterministic}
    \hat{d}_S(t_3,t_2,t_1)= \sum_{m=1}^{12} \sum_{d=1}^{7} \Big( \beta_{S,0,m,d}+\sum_{h=1}^{23}\beta_{S,h,m,d}~ToH_{h}(t_3,t_2, t_1)\Big)
\end{equation}
Where $\beta_{S,0,m,d}$ refers to the intercept of each model and $\beta_{S,h,m,d}$ are twenty-three regression coefficients for each month (m) and weekday (d) combination respectively.

Given the nature of the linear regressive component defined in Equation \ref{eq:shortterm_deterministic}, it is obvious that the residuals inhabit different characteristics depending on the month and day. Therefore, a different AR model is fitted to the residuals of each month. The 12 respective AR models are determined in a similar way as described in Section \ref{sec:Medium_stochastic}. As the residual time series of all 12 months are stationary, an ARMA as opposed to an ARIMA model approach is used. The results of the ACF suggested that the residuals display seasonal auto-correlation. Because of the periodic nature of 24-hours in a day, initially a seasonal ARMA model with seasonality at lag 24 was implemented. This approach failed to fully resolve the auto-correlation in the residuals. Therefore, the lowest p and q orders - for which no significant auto-correlation is left - are determined and the respective ARMA models are calculated. For most months, this resulted in an ARMA(27,26) model. Contrary to the mid-term linear regressive residuals, the short-term model residuals showed a more prominent recurrent pattern. Thus, the ARMA predictions for which the forecast horizon is still within the lag order of the model, can be used for the forecast of all consecutive days within that month as well. Because this approach demonstrates very good results, another LSTM approach for the short-term residuals isn't necessary.

\begin{figure}[ht]
\centering
\begin{subfigure}{.5\textwidth}
  \centering
  \includegraphics[width=0.9\linewidth]{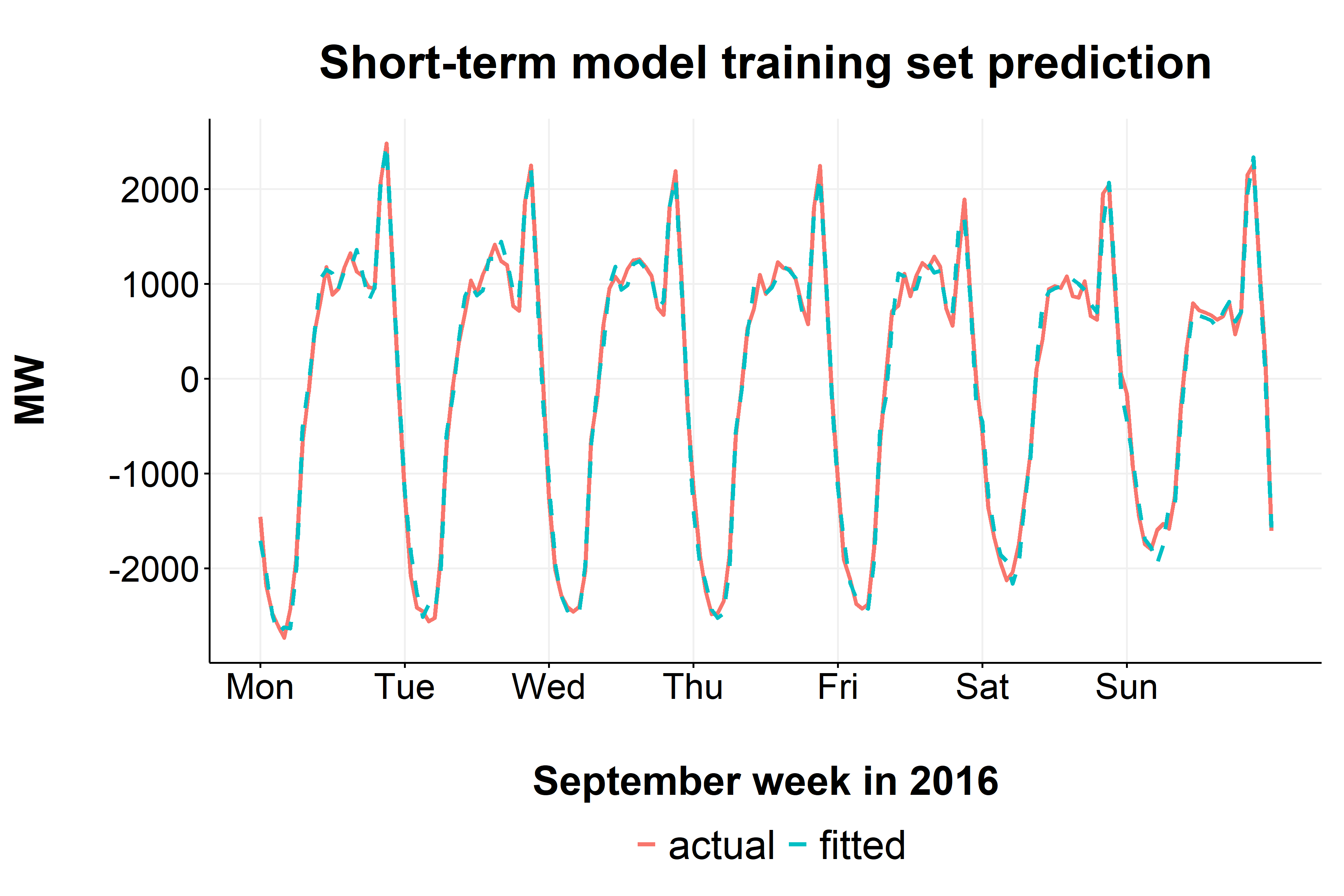}
  \caption{In-sample short-term predictions.}
  \label{fig:ST_training_full}
\end{subfigure}%
\begin{subfigure}{.5\textwidth}
  \centering
  \includegraphics[width=0.9\linewidth]{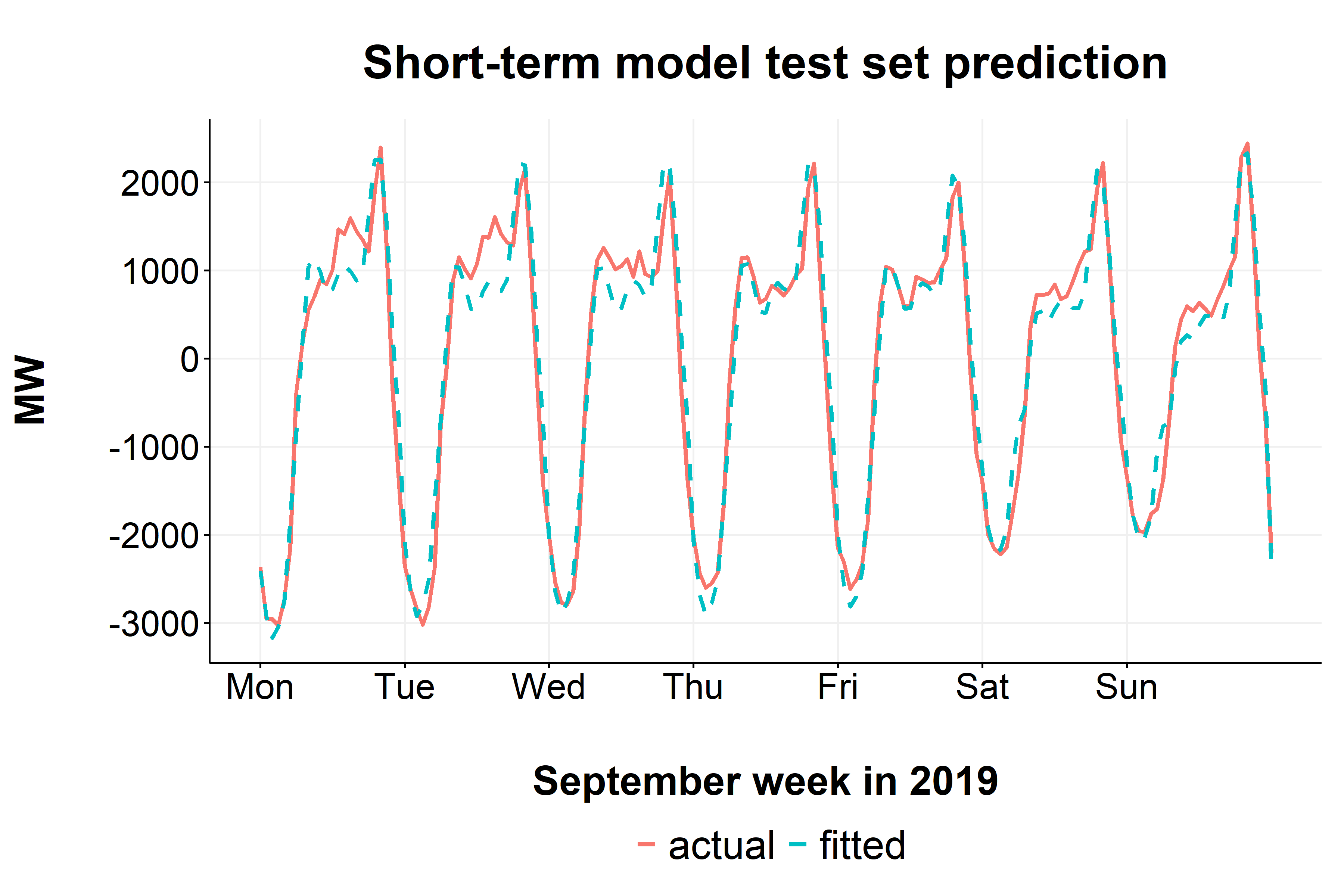}
  \caption{Out-of-sample short-term predictions}
  \label{fig:ST_testset_2019}
\end{subfigure}
\caption{Short-term load pattern and model forecast for a sample week within the training and test set.}
\label{fig:ST_det_predictions_test}
\end{figure}

Figure \ref{fig:ST_training_full} shows the in-sample predictions for a week in September in 2016, whereas Figure \ref{fig:ST_testset_2019} shows the out-of-sample forecasts for the same week in 2019.
A closer examination of Figure \ref{fig:ST_det_predictions_test} reveals that the seasonal daily load pattern wasn't subjected to much change over the course of the years from 2016 to 2019 and is described very well by our chosen CST approach.

\section{Forecasting Results}

We combined and evaluated all the model components defined in the additive Equation \ref{eq:complete_model} to forecast electricity demand in the test periods 2019 and 2020.  Table \ref{tab:forecast_errors_cst} 
shows four different error parameters (RMSE, MAPE, MAE, MASE) for the respective forecast horizons. The forecast including the training set values from 2013 to 2018 is reported as well to provide an overview of the in-sample accuracy as well. The RMSE and MAE are reported in megawatt (MW) with hourly demand values in 2019 and 2020 ranging from 10905 MW/h to 23642 MW/h. The increase in MAPE for the forecast of 2020 compared to 2019 can likely be contributed to the start of the COVID-19 global pandemic. Although the forecasting accuracy declines slightly in 2020, a MAPE around 3.17 \% is still a very good result, particularly for a forecast horizon of 17520 predicted hourly values.      

\begin{table}[H]

\ra{1.3}
  \begin{center}
    
    \caption{Comparison of stochastic midterm residual prediction.}
    \label{tab:forecast_errors_cst}
    \begin{tabular}{@{}l|cccc@{}} 
      \toprule 
       \textbf{}&\textbf{RMSE} & \textbf{MAPE} & \textbf{MAE}  & \textbf{MASE}\\
      \textbf{}&\footnotesize{[MW]} & \footnotesize{[\%]} & \footnotesize{[MW]}  & \textbf{}\\
      \midrule
2013 to 2020   &  599.1 & 2.78 & 473.2 & 0.372 \\
forecast 2019 \& 2020    &  682.4 & 3.17 & 516.5 & 0.457\\
forecast 2019 &        619.8 & 2.79 & 467.8 &   0.473\\
forecast 2020 &        739.6 & 3.56 & 565.3 & 0.444\\
      \bottomrule
    \end{tabular}
  \end{center}
\end{table}

\begin{figure}[ht]
    \centering
    \includegraphics[width=0.8\textwidth]{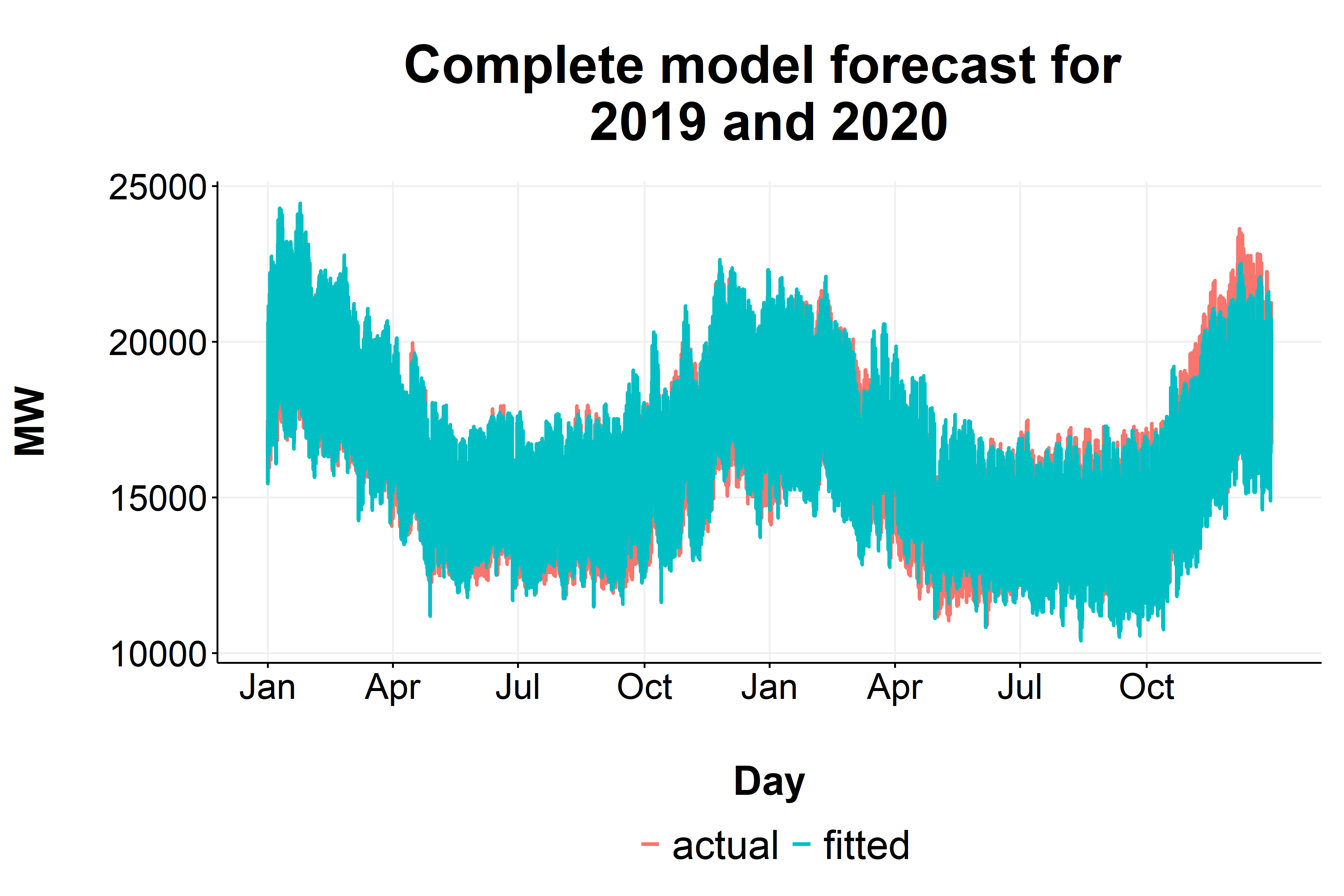}
    \caption{Complete forecast of Ukrainien hourly demand over the test period of 2019 and 2020.}
    \label{fig:full_model}
\end{figure}

\begin{figure}[ht]
\centering
\begin{subfigure}{.5\textwidth}
  \centering
  \includegraphics[width=0.9\linewidth]{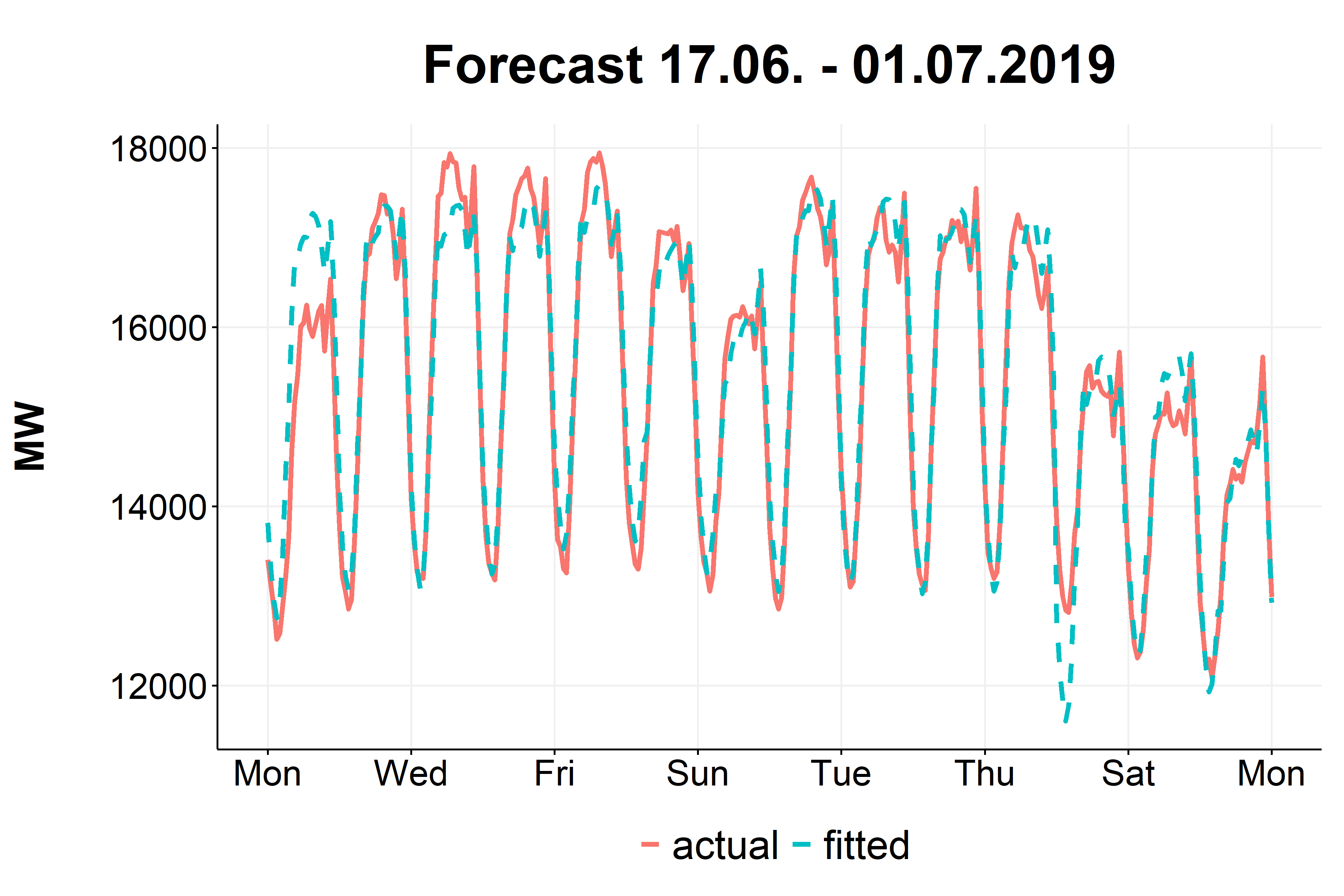}
  \caption{Two sample weeks in June 2019.}
  \label{fig:2weeks_2019}
\end{subfigure}%
\begin{subfigure}{.5\textwidth}
  \centering
  \includegraphics[width=0.9\linewidth]{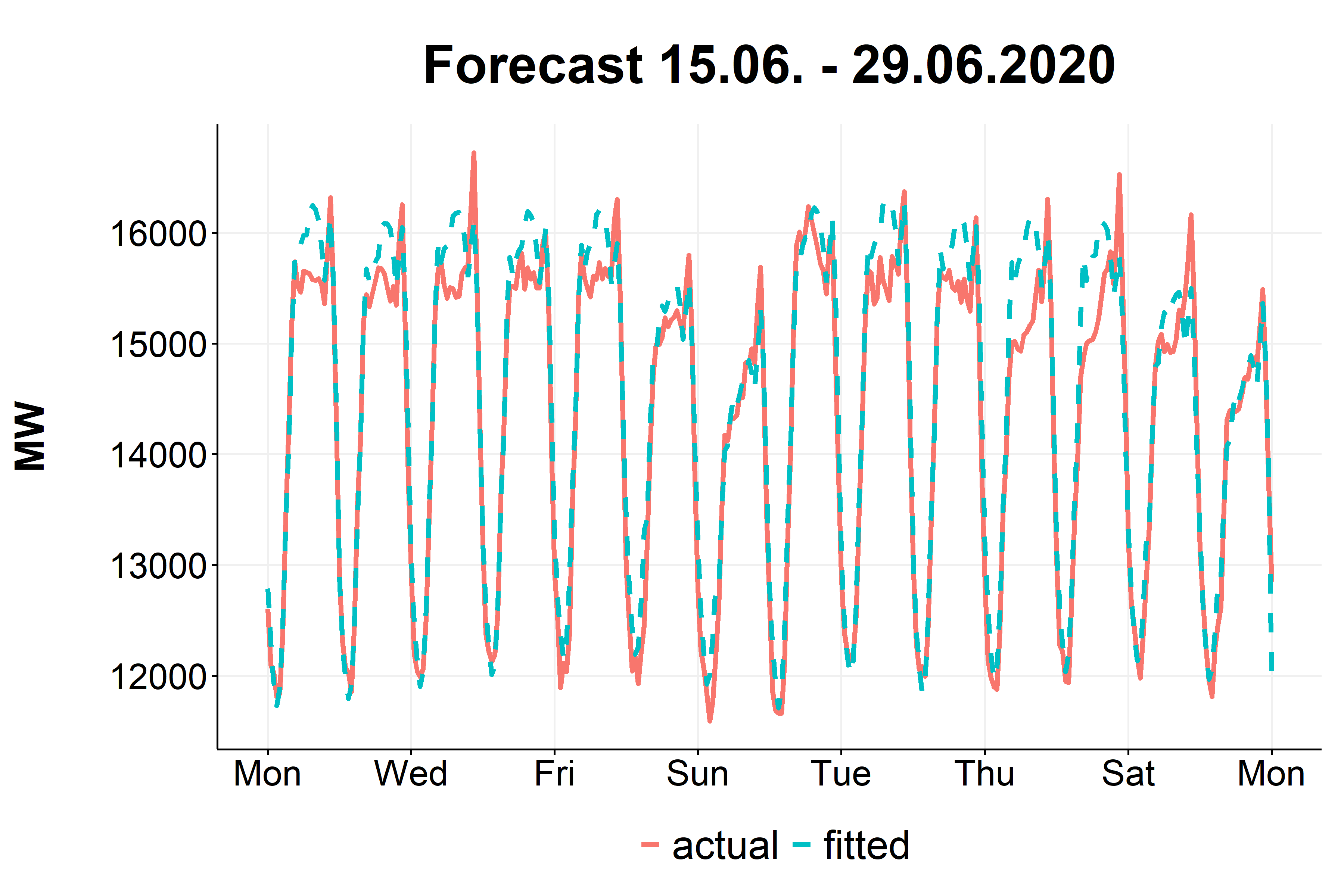}
  \caption{Two sample weeks in June 2020.}
  \label{fig:2weeks_2020}
\end{subfigure}
\caption{Forecast and actual demand of two sample weeks in June 2019 and June 2020, starting from Monday.}
\label{fig:fullmodel_weeks}
\end{figure}

Figure \ref{fig:full_model} shows the complete model forecasts and actual values for 2019 and 2020. Two sample weeks in the middle of 2019 and 2020 respectively are displayed in Figure \ref{fig:fullmodel_weeks} to provide more detail about the forecasted and actual demand structure, as well as typical error occurrences. In Figure \ref{fig:2weeks_2019}, it can be observed that the first Monday (17.06.) shows an atypically low demand over the daylight hours that is not captured well by the forecast as well as an underestimation of load for the night of Friday (28.06.) to Saturday (29.06.). Apart from that, the forecasted demand structure seems to be well in line with the actual demand for that time period. The same weeks in 2020 (Figure \ref{fig:2weeks_2020}) show a more stable structure that is slightly overestimated in the noon hours by the forecast, but is otherwise captured quite well.

\section{Conclusion}
This article presents a novel hybrid model for national electricity demand prediction. Input data includes Ukraine's hourly electricity demand, macroeconomic variables, and temperature from 2013 to 2020. The proposed methodology 
first decomposes the data into three categories with respect to time scale: a long-term model covering yearly forecasts, a medium-term model forecasting days ahead, and a short-term model covering forecasts in an hourly resolution. The long-term model is well captured through a linear regression model on macroeconomic and demographic variables. The mid-term model integrates temperature and calendar regressors to describe the underlying structure, and ARIMA and LSTM ``black-box'' pattern-based approaches to describe the residual. Lastly, the short-term part is well captured through linear regressors on calendar variables and different ARMA models for the residual. The linear regressive and ARIMA models were calculated in R \cite{R}. The LSTM model was written in Python \cite{Python}. Both codes are available as open source software at \cite{github22}. 

The main strengths of the proposed hybrid algorithm are two-fold. On the one hand, it describes the underlying structure and relationships in the data through classical statistical regression models. This allows for understanding of current and future drivers that influence electricity demand in different time horizons. Knowledge on influential external factors allows decision-makers to react when the future underlying structure changes. On the other hand, it combines ARIMA and LSTM approaches to correct residuals and increase forecast accuracy. 
The main drawback of this model is that for future predictions, high-resolution forecasts of macroeconomic data is required. Also, historical weather data is projected for the future. 

Future research should consider reproducing and reusing this algorithm for other developing and industrialized countries. To facilitate this task the methodology has been developed with an option for automation in mind. All necessary data can be acquired using automatic API requests and the model calculation and evaluation can be fully automated as well.

\bigskip
\bigskip



\bibliographystyle{model1-num-names}
\bibliography{Literature.bib}







\end{document}